\documentclass[10pt,twocolumn,letterpaper]{article}

\usepackage{wacv}
\usepackage{times}
\usepackage{epsfig}
\usepackage{graphicx}
\usepackage{amsmath}
\usepackage{amssymb}

\usepackage{multirow}
\usepackage{gensymb}
\usepackage{subcaption}
\usepackage[labelsep=period]{caption}
\usepackage{siunitx}
\usepackage[page]{appendix}

\DeclareMathOperator*{\argmin}{arg\,min}

\usepackage[pagebackref=true,breaklinks=true,letterpaper=true,colorlinks,bookmarks=false]{hyperref}

\wacvfinalcopy 

\def\wacvPaperID{295} 

\begin{document}

\title{Audio--Visual Model Distillation Using Acoustic Images}

\author{Andr\'{e}s F. P\'{e}rez\textsuperscript{1}\qquad Valentina Sanguineti\textsuperscript{1,2}\qquad Pietro Morerio\textsuperscript{1}\qquad Vittorio Murino\textsuperscript{1,3,4}\\
{\small \textsuperscript{1}Pattern Analysis \& Computer Vision - Istituto Italiano di Tecnologia, }
{\small \textsuperscript{2}Universit\`{a} degli Studi di Genova, Italy, }\\
{\small \textsuperscript{3}Computer Science Department - Universit\`a di Verona, Italy, }
{\small \textsuperscript{4}Huawei Technologies Ltd., Ireland Research Center}\\
{\tt\small andres.perez@mail.polimi.it \{valentina.sanguineti, pietro.morerio, vittorio.murino\}@iit.it}
}

\maketitle
\ifwacvfinal\thispagestyle{empty}\fi

\begin{abstract}
In this paper, we investigate how to learn rich and robust feature representations for audio classification from visual data and acoustic images, a novel audio data modality.
Former models learn audio representations from raw signals or spectral data acquired by a single microphone, with remarkable results in classification and retrieval. However, such representations are not so robust towards
variable environmental sound conditions.
We tackle this drawback by exploiting a new multimodal labeled action recognition dataset acquired by a hybrid audio-visual sensor that provides RGB video, raw audio signals, and spatialized acoustic data, also known as acoustic images, where the visual and acoustic images are aligned in space and synchronized in time.
Using this richer information, we train audio deep learning models in a teacher-student fashion. In particular, we  distill knowledge into audio networks from both visual and acoustic image teachers.
Our experiments suggest that the learned representations are more powerful and have better generalization capabilities than the features learned from models trained using just single-microphone audio data.
\end{abstract}

\section{Introduction}


Humans experience the world through a number of simultaneous sensory observation streams. The co-occurrence of these streams provides a useful learning signal to understand the environment surrounding us \cite{Gaver1993}. There is in fact evidence that audio-visual mirror neurons play a central role in the recognition of actions given their temporal synchronization \cite{Arrighi2009}. Furthermore, it was found that many neurons with receptive fields spatially aligned across modalities show a super-additive response to coincident and co-localized multimodal stimulations \cite{Wallace1993}.

\begin{figure}[t]
\centering
\includegraphics[width=\linewidth]{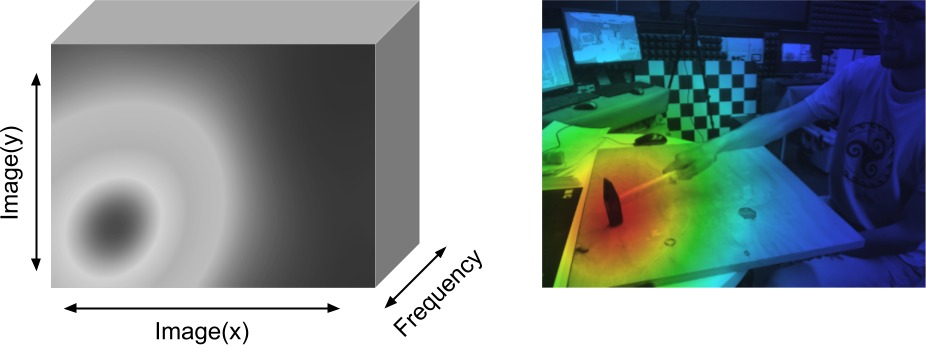}
\caption{{\small \textit{Left}: multispectral acoustic image volume associated to the audio content of the sensed scene. It has two spatial dimensions (aligned with the visual image space) and a frequency axis of 512 bins that cover the sensor's audible range. Each image in the volume represents the spatial audio information associated to each frequency bin. \textit{Right}: visualization (as heat color map) of an acoustic image formed by summing the energy of every frequency bin between $900 Hz$ and $6400 Hz$ for each spatial location, overlaid on the corresponding video frame.
The spatial location of the audio signal with the highest intensity is identified in red.
}}
\label{fig:acoustic_image}
\vspace{-15pt}
\end{figure}

In this paper, motivated by these findings, we investigate whether and how visual and acoustic data \textit{synchronized in time} and \textit{aligned in space} can be exploited for scene understanding.  
We take advantage of a recent audio-visual sensor, called DualCam, composed by an optical camera and a 2D planar array of microphones (see Figure~\ref{fig:dualcam}), able to provide spatially localized acoustic data aligned with the corresponding optical image (see  Figure~\ref{fig:acoustic_image}, right) \cite{Zunino2015}. Specifically, by combining the raw signals acquired by 128 microphones (by beamforming \cite{VanTrees2002}), this sensor is able to output an acoustic image where each pixel represents the imprint of the sound coming from the corresponding pixel location in the optical image. 
Using this sensor, we generate a new multimodal dataset depicting different subjects performing several actions in multiple scenarios. 
By exploiting spatialized audio information coupled to the related visual data and designing suitable multimodal deep learning models, we aim at generating more discriminant and robust features, likely resulting in a better description of the scene content for robust audio classification.  
Figure~\ref{fig:acoustic_image} shows the multispectral acoustic image used as input data, which has 512 frequency bins and an example of visualization of an acoustic image overlaid upon an optical image. 
%
%
%

The idea of leveraging the co-occurrence of visual and audio events as supervisory signal is not new. Former approaches in the pre deep-learning era combined visual and auditory signals in rather simplistic 
ways. 
For instance, in \cite{Yuhas1989} a neural network was trained to predict the auditory signal given the visual input. A particularly relevant earlier work is \cite{DeSa1993}, which introduced a self-supervised learning algorithm for jointly training audio and visual networks by minimizing codebook disagreement. Another interesting work is \cite{Kidron2005}, which presented an algorithm based on canonical correlation analysis (CCA) to detect pixels associated to the sound, while filtering out other dynamic (but silent) pixels.

Several recent works address audio-related tasks such as natural sound recognition \cite{Li2019}, speech separation and enhancement \cite{Afouras2018, Gabbay2017}, audio event classification or sound source localization \cite{Parekh2018, Tian2018}, either by directly modeling raw audio signals with 1D convolutions \cite{Aytar2016, Owens2018, Ravanelli2018} or, most popularly, by modeling intermediate sound representations such as spectrograms or cochleograms \cite{Arandjelovic2017, Arandjelovic2018, Owens2016, Owens2016a, Senocak2018, Senocak2018a, Takahashi2017, Zhao2018}.
Nevertheless, none of the past works tried to exploit spatially localized acoustic data to assess the potentialities of such richer information source. 

In our work, we claim that it is possible to train audio deep learning models to face an action recognition problem in a more robust way across different scenarios utilizing a teacher-student framework able to distill knowledge \cite{Garcia2018, Lopez-Paz2015}  from state-of-the-art vision network models and from a novel architecture that operates on the spatialized acoustic data.
Similarly to \cite{Ngiam2011}, our intuition is to learn better features for a given modality assuming the availability of other complementary modalities at training time. 
We leverage video and multispectral acoustic image sequences  aligned in space/time as side information at training, and predict actions given only a raw audio signal acquired by a single microphone at test, in a cross-scenario setting, where the environmental noise conditions are significantly different. Current methods, even best deep learning models, lead to very low classification accuracies \cite{Gharib2018, Mesaros2018} in such conditions.

Hence, in essence, in this work we try to answer the following question: \textit{Does spatialized data allow to learn more discriminant features for  single-microphone audio classification?}
In this respect, our  main contributions can be  summarized as follows.
\begin{enumerate}
\itemsep0em
\item We propose a thorough study to assess whether visual and acoustic data aligned in space and synchronized in time bring advantage for single-microphone audio classification.
\item We introduce a new multimodal dataset consisting in 14 action classes, in which acoustic and visual data are spatially aligned. This type of multi-sensory data has no counterpart in the literature and may lead to further studies by the scientific community.
\item We develop a deep teacher-student model to deal with such new data, 
showing that it is indeed possible to extract semantically richer representations for improving audio classification from single microphone. 
In particular, we distill knowledge learned from spatialized audio-visual modalities to a single-microphone model.
\end{enumerate}

It is worth to note that we are the first to propose an algorithm in which the transfer of knowledge involves teacher models considering 2 different modalities (2D audio and 2D visual data) and the student model is devised for a different modality (1D audio signal), when typically the student deals with the task of one of the teacher models.

We validate our approach 1) on the proposed action dataset, and 2) by transferring learned representations on a standard sound classification benchmark dataset, demonstrating remarkable capabilities and the usefulness of distillation for cross-scenario learning.

The remainder of this paper is organized as follows.
We first discuss the related work in Section~\ref{section:related_work}, mainly focusing on audio-visual models and benchmark datasets. 
In Section~\ref{section:dataset}, we describe our new acquired multimodal action dataset, and in Section~\ref{section:acoustic_model}, we describe acoustic image pre-processing and we propose the network architecture to deal with acoustic images.
In Section~\ref{section:model_training}, we present our distillation-based approach to deal with multispectral acoustic data, and in Section~\ref{section:experiments}, we extensively validate our proposed framework by devising a set of experiments in order to assess the soundness of the learned representations. Finally, we draw conclusions in Section~\ref{section:conclusions}.

\section{Related Work}\label{section:related_work}

We briefly review related work in the areas of multimodal learning, video and sound self-supervision, and transfer learning. We also review already existing audio and audio-visual datasets.





\textbf{Multimodal learning.}
Multimodal learning concerns relating information from multiple data modalities. Such data provides complementary semantic information due to correlations in between them \cite{Ngiam2011}. We consider the cross-modality learning setting, in which data from multiple modalities is available only during training, while only data from a single modality is provided at testing phase.
In \cite{Aytar2017, Castrejon2016} the authors learn shared representations from multimodal aligned data and use them for cross-modal retrieval. \cite{Aytar2017} for instance considers three major natural modalities: vision, sound and language, while \cite{Castrejon2016} considers five weakly aligned modalities: natural images, sketches, clip art, spatial text, and descriptions. Other works such as \cite{Garcia2018, Hoffman2016} utilize RGB video images and depth information to learn feature representations through modality hallucination. In our work instead, we consider RGB video images, raw audio and acoustic images for training phase, and only raw audio at testing time.

\textbf{Video and sound self-supervision.}
There has been increased interest in using deep learning models for multimodal fusion of auditory and visual signals to improve the performance of visual models or solve various speech-related problems, such as speech separation and enhancement.

First approaches trained single networks on one modality using the other one to derive some sort of supervisory signal \cite{Aytar2016, Harwath2016, Owens2016, Owens2016a, Owens2018a}. For example \cite{Aytar2016, Harwath2016} train an audio network to correlate with visual outputs using pre-trained visual networks as a teacher. Others such as \cite{Owens2016a, Owens2016} train a visual network to generate sounds by solving a regression problem consisting in mapping a sequence of video frames to a sequence of audio features. In \cite{Owens2018a} instead, they learn visual models using ambient sounds as scene labels.

More recent works \cite{Arandjelovic2017, Arandjelovic2018, Ephrat2018, Owens2018, Senocak2018} train both visual and audio networks aiming at learning multimodal representations useful for many applications, such as cross-modal retrieval, speech separation, sound source localization, action recognition, and on/off-screen audio source separation.
For instance in \cite{Arandjelovic2017, Arandjelovic2018} they learn aligned audio-visual representations, using an audio-visual correspondence task.
In \cite{Owens2018} they train an early-fusion multisensory network to predict whether video frames and audio are temporally aligned.
In \cite{Senocak2018} a two-stream network structure is trained utilizing an attention mechanism guided by sound information to localize the sound source.

They key factor in all these works is that they exploit the natural synchronization between auditory and visual signals by training in a self-supervised manner. Although we address our problem in a pseudo-supervised manner using a combination of hard and soft labels, we notice that the natural spatial alignment and time synchronization of the data produced by the DualCam sensor opens the door to also train models through self-supervision.

\textbf{Transfer learning.}
Our work is strongly related to transfer learning which deals with sharing information from one task to another. In particular we transfer knowledge between networks operating on different data modalities (see Section~\ref{section:model_training}). We perform transferring with the aid of the generalized distillation framework which proposes to use the teacher-student approach from the distillation theory to extract knowledge from a privileged information source \cite{Lopez-Paz2015}, also called a teacher. In our case the privileged information leveraged at training time is represented by the additional modalities, i.e. video and acoustic images.
A rather simple transfer mechanism is that of \cite{Aytar2016} which proposes a teacher-student self-supervised training procedure based on the Kullback-Leibler divergence to transfer knowledge from a vision model into sound modality using unlabeled video as a bridge. This mechanism resembles the generalized distillation framework, however they only rely on the teacher soft labels which are in general less reliable than hard labels.
An interesting work is \cite{Hoffman2016} which introduces a novel technique for incorporating additional information, in the form of depth images, at training time to improve test time RGB only detection models.
We draw inspiration from \cite{Garcia2018} which addresses action recognition by distilling knowledge from a depth network into a vision network. They accomplish this by training a hallucination network \cite{Hoffman2016} that learns to distill depth features. It is worth noticing that although \cite{Garcia2018} works with different data modalities, it is the closest to ours since they transfer knowledge with the aid of the generalized distillation framework.



\textbf{Audio-visual datasets.}
Due to recent interest in audio-visual and multimodal learning, several audio and audio-visual datasets have emerged. Here we summarize some of the most prominent ones.

\textit{Flickr-SoundNet} \cite{Aytar2016} is a large unlabelled dataset of completely unconstrained videos from Flickr, compiled by searching for popular tags and dictionary words. It contains over 2 million videos which total for over one year of continuous natural sound and video.

\textit{Kinetics-Sounds} \cite{Arandjelovic2017} comprises a subset of the Kinetics dataset \cite{Kay2017}, which contains YouTube videos manually annotated for human actions, and cropped to 10 seconds around the action.
The subset contains 19k video clips formed by filtering the Kinetics dataset for 34 human action classes, which have been chosen to be potentially manifested visually and aurally.

\textit{FAIR-Play} \cite{Gao2018a} is an unlabelled video dataset with binaural audio that mimics human hearing. It consists of 1.871 short clips of 10 seconds long musical performances, totaling 5.2 hours. It depicts different combinations of people playing musical instruments including cello, guitar, drum, ukelele, harp, piano, trumpet, upright bass, and banjo, in a large music room, in solo, duet, and multiplayer performances.

\textit{Environmental Sound Classification (ESC-50)} \cite{Piczak2015} is a labeled collection of 2.000 environmental audio recordings manually extracted from Freesound. It consists of 5 seconds long recordings organized into 50 semantical classes loosely arranged into five major categories: animals, natural soundscapes \& water sounds, human non-speech sounds, interior/domestic sounds, and exterior/urban noise.

\textit{Detection and Classification of Acoustic Scenes and Events (DCASE)} \cite{Mesaros2018} is a dataset consistent of of recordings from various acoustic scenes. It was recorded in six large European cities, in different locations for each scene class. For each recording location there are 5 to 6 minutes of audio split into segments of 10 seconds.

The closest dataset to ours is \textit{FAIR-Play} because of its size and the nature of its data, since binaural audio is a form of spatial audio. Similarly to \textit{Kinetics-Sounds} we propose a dataset of human actions, but with data in multiple modalities which try to describe more realistic conditions.

\section{Audio-Visually Indicated Action Dataset}\label{section:dataset}

We introduce a new multimodal dataset comprised of visual data as RGB image sequences and acoustic data as raw audio signals acquired from 128 microphones. The latter signals, opportunely combined by a beamforming algorithm, compose a  multispectral acoustic image volume, which is aligned in space and time with the optical images (see Figure~\ref{fig:acoustic_image}). 
The following 14 actions were chosen:

\begin{enumerate}
\itemsep0em 
    \begin{minipage}{0.4\linewidth}
        \item Clapping
        \item Snapping fingers
        \item Speaking
        \item Whistling
        \item Playing kendama
        \item Clicking
        \item Typing
    \end{minipage}
    \hfill
    \begin{minipage}{0.4\linewidth}
        \item Knocking
        \item Hammering
        \item Peanut breaking
        \item Paper ripping
        \item Plastic crumpling
        \item Paper shaking
        \item Stick dropping
    \end{minipage}
\end{enumerate}

\begin{figure}[h]
  \begin{subfigure}[t]{.32\linewidth}
    \centering\includegraphics[width=\linewidth]{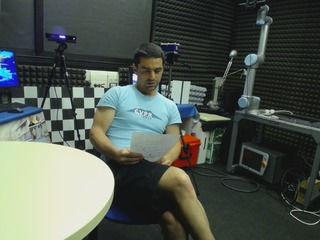}\\
    \centering\includegraphics[width=\linewidth]{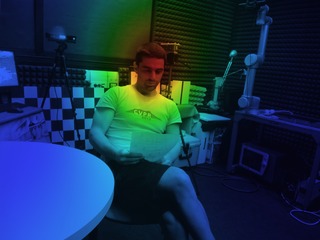}
    \\
   \centering\includegraphics[width=\linewidth]{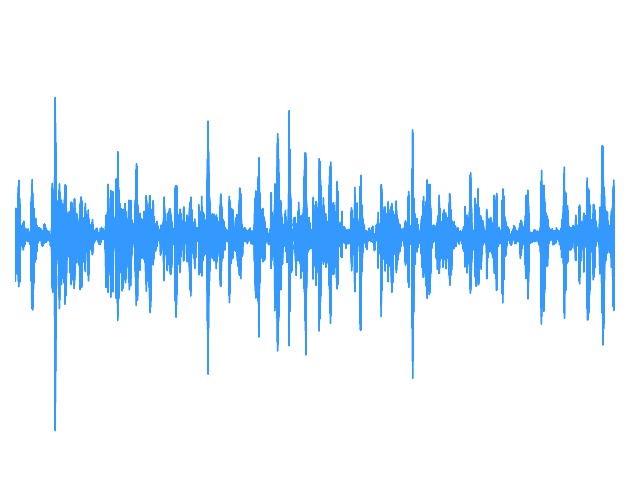}
   \vspace{-1cm} \caption{}\label{fig:dataset_location_1}
    
  \end{subfigure}
  \begin{subfigure}[t]{.32\linewidth}
    \centering\includegraphics[width=\linewidth]{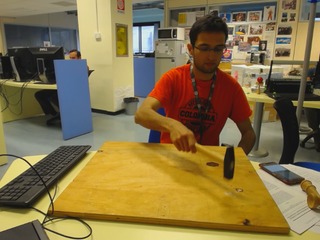}\\
    \centering\includegraphics[width=\linewidth]{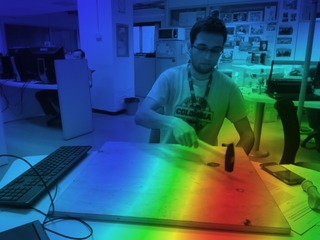}
    \\
    \centering\includegraphics[width=\linewidth]{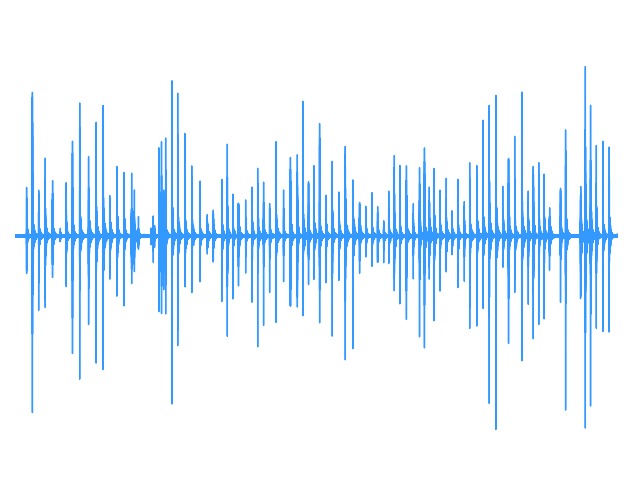}
   \vspace{-1cm} \caption{}\label{fig:dataset_location_2}
    
  \end{subfigure}
  \begin{subfigure}[t]{.32\linewidth}
    \centering\includegraphics[width=\linewidth]{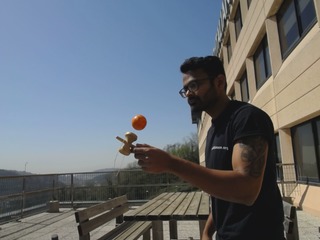}\\
    \centering\includegraphics[width=\linewidth]{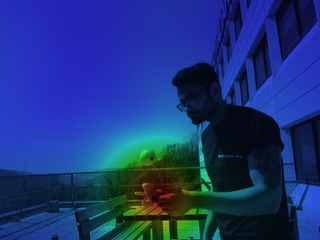}
    \\
    \centering\includegraphics[width=\linewidth]{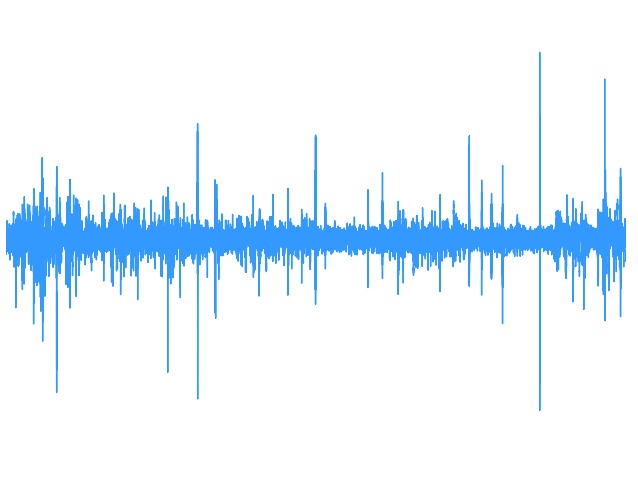}
   \vspace{-1cm} \caption{}\label{fig:dataset_location_3}
    
  \end{subfigure}
  
  \vspace{-0.2cm} 
  \caption{{\small Three examples of Audio-Visual Indicated Actions dataset represented as video frame, acoustic image visualization overlaid on the frame, and raw waveform (from a single microphone). 
  (a) Speaking in anechoic room.
  (b) Hammering in the indoor open space area. 
  (c) Playing Kendama in the terrace. }}
  \label{fig:dataset}
\end{figure}

For the acquisition, we acknowledge the participation of 9 people performing the aforementioned actions recorded in three different scenarios, with increasing and varying noise conditions, namely, an anechoic room, an indoor open space area, and a terrace outdoor. 
We name them  scenario 1, 2, and 3, respectively. In our dataset, the same action is performed by different subjects in distinct places, so allowing to show the equivariance properties of the multispectral acoustic images across subjects, scenarios and position in the scene, which are exploited when learning audio features from an acoustic teacher model.
In the end, the dataset consists of 378 audio-visual video sequences (27 per action) between 30 and 60 seconds depicting different people individually performing a set of actions producing a characteristic sound in each scenario.
Figure~\ref{fig:dataset} shows representative samples of our dataset for the 3 considered scenarios.

\begin{figure}[h]
  \centering
  \includegraphics[width=0.32\linewidth]{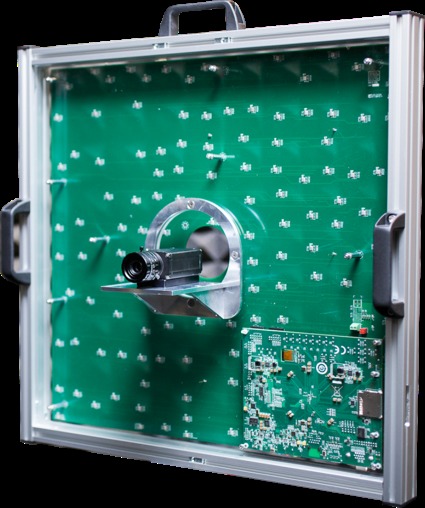}
  \caption{{\small DualCam acoustic-optical camera.}}
  \label{fig:dualcam}
\end{figure}

We acquired the dataset using the DualCam acoustic-optical camera described in \cite{Zunino2015}. The sensor captures both audio and video data using a $0.45m\times0.45m$ planar array of 128 low-cost digital MEMS microphones located according to an optimized aperiodic layout, and a video camera placed at the device center as depicted in Figure~\ref{fig:dualcam}.

The device is capable of acquiring audio data in the range \SI{200}{\Hz} -- \SI{10}{\kHz} and audio-video sequences at a frame rate of 12 fps.
In our acquisition setup the camera was static looking at the scene, while the subjects moved around within its field of view
at a minimum distance of 2 meters from the device.

After collecting the dataset, audio and video data had to be synchronized since they were acquired in an interleaved way at different frame rates. 

The data provided by the sensor consists in RGB video frames of $640\times480$ pixels, raw audio data from 128 microphones acquired at a frequency of 12 kHz, and $36\times48\times512$ multispectral acoustic images obtained from the raw audio signals of all the microphones using beamforming, which summarize the per-direction audio information in the frequency domain. 
This means that each acoustic pixel corresponds to 13,3 visual pixels, in fact acoustic resolution is lower than optical one. 
Among the raw audio waveforms, we choose the one of just one microphone for testing single microphone audio networks.

\section{Learning with Acoustic Images}\label{section:acoustic_model}

In this section, we describe acoustic images representation, their pre-processing and the network architecture we proposed for modelling this novel type of data.

%

\textbf{Acoustic Images Pre-processing.} Multispectral acoustic images are generated with the frequency implementation of the filter-and-sum beamforming algorithm \cite{VanTrees2002}, aimed at producing a volume of size $36\times48$, with 512 channels corresponding to the frequency bins which represent the frequency information. 
Full details of the algorithm can be found in \cite{Zunino2015}.

Handling input acoustic images with 512 channels is a computationally expensive task and typically the majority of information in our dataset is contained in the low frequencies. Consequently, we decided to compress the acoustic images using Mel-Frequency Cepstral Coefficients (MFCC), which consider audio human perception characteristics \cite{Terasawa2006}. 
Therefore, we compute 12 MFCC, going from from $36\times48\times512$-D volumes to $36\times48\times12$-D volumes, retaining the most important information and reducing consistently the computational complexity and the memory footprint.

\textbf{DualCamNet Architecture.} 
Acoustic images provide a small temporal support which is generally not enough for discriminating information over time intervals of several seconds. For this reason, we feed to our network a set of 12 consecutive $36\times48\times12$ acoustic images corresponding to 1 second of audio data.
We deem that 1 second of acoustic images is a reasonable trade-off between sound information content and processing cost.

\begin{figure}[htbp]
  \centering
  \begin{subfigure}[t]{.25\linewidth}
    \includegraphics[width=\linewidth]{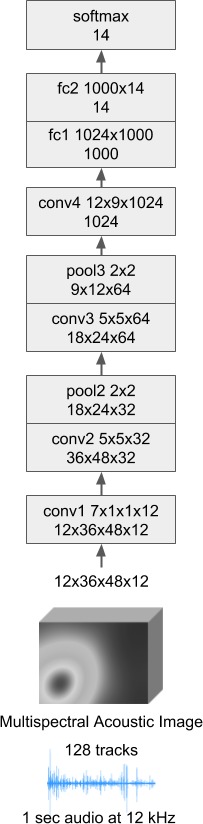}
    \caption{DualCamNet}
    \label{fig:network_dualcamnet}
  \end{subfigure}
  \hspace{0.1\linewidth}
  \begin{subfigure}[t]{.25\linewidth}
    \includegraphics[width=\linewidth]{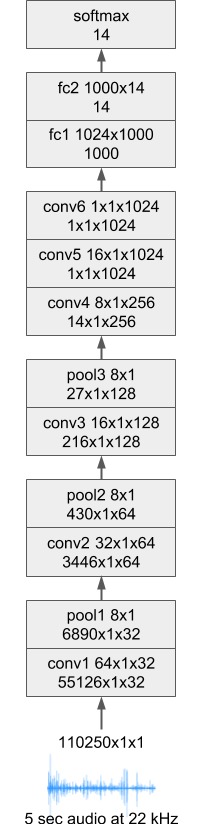}
    \caption{\small OursSoundNet}
    \label{fig:network_soundnet}
  \end{subfigure}
  \hspace{0.1\linewidth}
  \begin{subfigure}[t]{.25\linewidth}
    \includegraphics[width=\linewidth]{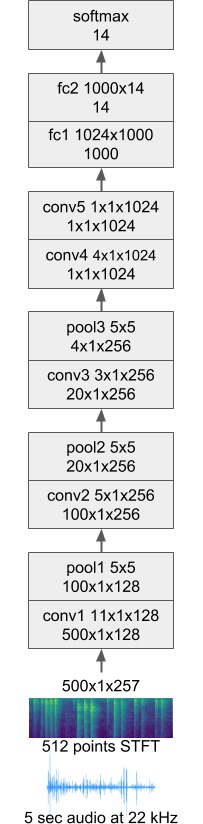}
    \caption{HearNet}
    \label{fig:network_hearnet}
  \end{subfigure}
\caption{{\small Our proposed networks. (a) DualCamNet architecture, used as teacher model. (b) OursSoundNet architecture,  used as student model. (c) HearNet architecture, used as student model.}}
\label{fig:network}
\end{figure}

In order to train a model able to discriminate information from acoustic images, we explicitly model both the spatial and the temporal relationships among them. To this end, we propose the architecture structure shown in Figure~\ref{fig:network_dualcamnet} which utilizes 3D convolutions as commonly done in visual action recognition \cite{Tran2018}, where the spatial and temporal convolutions are decoupled. 
%

We follow the LeNet \cite{LeCun1999} design style, with $5\times5$ convolutional filters, and $2\times2$ max-pooling layers with stride 1 and zero-padding to keep the spatial resolution. The network includes 3 blocks of convolutional layers plus a block of 3 fully convolutional layers which produces the output prediction.

The first block consists of a single 1D convolutional layer over time followed by a ReLU nonlinearity. The aim of this layer is modeling the temporal relationship of consecutive acoustic images by aggregating them. In particular, we apply a filter of size 7 with stride 1 and zero-padding to keep the temporal resolution. We experimented with several filters sizes finding 7 to be the best one.


The second and third blocks model the spatial equivariance of the acoustic images and consist of a 2D convolutional layer followed by max-pooling. We go from the 12 channels of the input to 32 channels and then double it to 64. Each convolutional layer is followed by batch normalization \cite{Ioffe2015} and ReLU nonlinearity.

The final block comprises 3 fully convolutional layers with ReLU in between. It converts the input feature map into a 14-D classification vector as output, namely the predicted class probabilities, using intermediate features size of 1024-D and 1000-D. 

This model will be used as teacher network in our validation experiments.

\section{Model Distillation}\label{section:model_training}

In this section, we describe the utilized network architectures and the knowledge transfer procedure.

\subsection{Architectures}\label{subsection:architectures}

Similarly to \cite{Garcia2018}, we utilize data from multiple modalities at training phase, and only data from a single modality at testing phase. We leverage
either RGB video images or multispectral acoustic images in training as side information, and we test only on audio data from a single microphone. 

We want to emphasize here that, to the best of our knowledge, this is the first time that model distillation is performed from modalities different from those utilized in testing. Specifically, we train on 2-dimensional spatialized audio and video data, to improve accuracy on a model working on mono-dimensional audio signals only as input. As further original aspect, \cite{Garcia2018} trains one ResNet-50 \cite{He2016} network per stream, while we use different network architectures for each stream of our model.

\textbf{Teacher Networks.} For the visual stream, we experimented with two models, ResNet-50 \cite{He2016} and its variation including 3D temporal convolutions introduced in \cite{Garcia2018}, here called Temporal ResNet-50. We choose ResNet-50 respect to Temporal ResNet-50 as it provides a good compromise between network size and accuracy.
On the other hand, Temporal ResNet-50 stands as a strong action recognition model dealing with  action dynamics with the aid of temporal connections between residual units.
It has also been selected since it constitutes a powerful baseline model to compare with.
DualCamNet, explained earlier in Section~\ref{section:acoustic_model}, will be used as teacher model as well in the following.

\textbf{Student Networks.} Regarding the raw audio waveform stream, we experimented two models that capture different characteristics of audio data. The first one is SoundNet \cite{Aytar2016}, which operates over time domain signals. We preferred the 5-layer version over the 8-layer one, as our dataset is not big enough to allow SoundNet to grasp the underlying data patterns. We used the exact same architecture described in \cite{Aytar2017}, adding 3 fully convolutional layers at the bottom of the network with 1024, 1000 and 14 filters, respectively. To avoid further confusion, we named our version \textit{OurSoundNet}.

The second model is a network based on the sound sub-network presented in \cite{Aytar2017}, called from here on, \textit{HearNet}.
Its architecture is shown in Figure~\ref{fig:network_hearnet}.
This network operates on amplitude spectrograms obtained from an audio waveform of 5 seconds, upsampled to \SI{22}{kHz}. Such spectrogram was produced by computing the STFT \footnote{Short-Time Fourier Transform} considering a window length of \SI{20}{\milli\second} with half-window overlap. This produces 500 windows with 257 frequency bands. The resulting $500\times1\times257$ spectrogram is interpreted as a 257-dimensional signal over 500 time steps.

HearNet processes spectrograms with 3 1D convolutions using kernel sizes 11, 5, 3, and 128, 256, 256 filters, respectively, with stride 1. The last convolutional layers are fully convolutional and use 1024, 1024, 1000 and 14 filters to obtain the class predictions. We applied zero-padding in all layers except conv4 in order to keep the spatial resolution. The chosen activation function is ReLU. After each of the first 3 convolutional layers, we downsampled with one-dimensional max-pooling by a factor of 5.

\subsection{Training procedure}

\begin{figure}[htbp]
  \centering
  \includegraphics[width=0.9\linewidth]{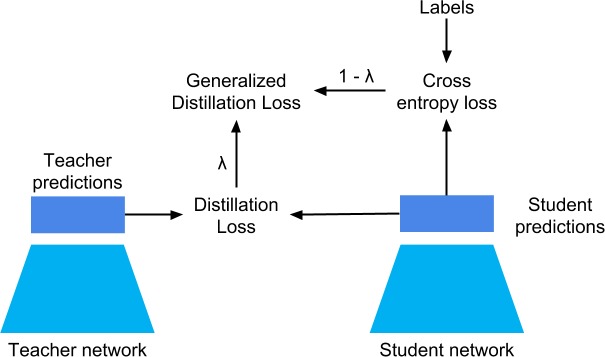}
  \caption{{\small Teacher-student training procedure}}
  \label{fig:teacher_student}
\end{figure}

Following the \textit{generalized distillation} framework \cite{Lopez-Paz2015}, we first learn a teacher function $f_t \in \mathcal{F}_t$ by solving a classification problem and, second, we compute the teacher soft labels $s_i$. As third step, we distill $f_t \in \mathcal{F}_t$ into $f_s \in \mathcal{F}_s$ by using both the hard and soft labels. The knowledge transfer procedure is graphically illustrated in Figure~\ref{fig:teacher_student}.

In particular we transfer knowledge between multiple modality network streams by using Hinton's distillation loss \cite{Hinton2015} to extract knowledge from privileged representations.
More formally, we distill the teacher learned representation $f_t \in \mathcal{F}_t$ into $f_s \in \mathcal{F}_s$ as follows:
\begin{equation}
\small
f_s = \argmin_{f \in \mathcal{F}_s} \frac{1}{n} \sum_{i=1}^n (1-\lambda)\ell(y_i, \sigma(f(x_i)))+\lambda\ell(s_i, \sigma(f(x_i))),
\end{equation}
where $ s_i = \sigma(f_t(x_i^*/T)) \in \delta^c $
are the soft labels derived from the teacher about the training data, $\mathcal{F}_t$ and $\mathcal{F}_s$ are classes of functions described by the teacher and student models \cite{Lopez-Paz2015}, respectively, $\sigma$ is the softmax operator, and  $y_i$ are the ground truth hard labels. 
The imitation parameter $\lambda \in [0,1]$ allows to balance the weight of soft labels with respect to the true hard labels $y_i$. 
The temperature parameter $T>0$ allows to smoothen the probability vector predicted by the teacher network $f_t$.

\section{Experimental Results}\label{section:experiments}

Our goal is to learn feature representations for raw audio data by transferring knowledge across networks operating on different data modalities. To evaluate how well our method addresses this problem we perform two sets of experiments with the objectives of 1) showing the improvement brought by distilling knowledge from different data modality networks and 2) assessing the quality of the distilled representations on a standard sound classification benchmark.

\subsection{Acoustic Features Transfer}
In this first set of experiments, we evaluate the performance of the teacher and student networks on the task of action recognition on our dataset. We train both the teacher and student networks in a fully supervised manner using action labels as ground truth, and only the student networks following the distillation procedure described in Section~\ref{section:model_training}. In all cases we trained for 100 epochs\footnote{The number of iterations varies with the size of the training set.} with batches of 32 elements using the Adam optimizer \cite{Kingma2014} with learning rates of \num{1e-3} and \num{1e-4} (see details in Supplementary Material). In order to measure the generalization capabilities of the learned representations, we evaluate the accuracy of our trained models on a cross-scenario setting, i.e. when the model is trained on certain scenario, it is tested on the other two scenarios using all the available data. In all cases the data was split by assigning 80\% of them for training, 10\% for validation and 10\% for test.

\textbf{Teachers Networks.} First, we train our DualCamNet model and the two proposed visual networks, ResNet-50 and Temporal ResNet-50, as they constitute our baselines. Table~\ref{table:teachers_accuracy} shows their performance. We observe that our DualCamNet convincingly outperforms the visual networks in all combination of scenarios. This indicates that most of the actions in our dataset are better distinguishable aurally than visually. One possible explanation for this, is that in the majority of the cases the ''object'' involved in the action execution, e.g. mouth, mouse or hammer, is not easily visible but has a characteristic sound signature.

A comparison of the two visual networks reveals that they achieve similar results throughout all configurations, indicating that motion is not a key factor to model the actions performed in our dataset. Consequently, we choose ResNet-50 over Temporal ResNet-50 as visual teacher for the rest of the experiments since the former one has a simpler structure.

Additionally, we have designed a hybrid network which combines the output of the DualCamNet and ResNet-50, to check whether modality fusion brings any performance improvement. We do so by concatenating the 1024 feature volumes of the two networks and processing them with two fully convolutional layers of 1000 and 14 filters, respectively. This network achieves a 7.1\% improvement in accuracy with respect to DualCamNet when trained over all scenarios.
It is important to note that it also consistently improves the testing accuracy in all cross-scenario configurations (see Table~\ref{table:teachers_accuracy}, AV column). These findings indicate some benefits brought by modality fusion that can be further explored in future research.

\begin{table}[h]
\centering
\footnotesize
\begin{tabular}{|l|l|c|c|c|c|}%
\hline
\multicolumn{1}{|c|}{\textbf{Train set}} & \multicolumn{1}{c|}{\textbf{Test set}} & \textbf{D} & \textbf{R} & \textbf{T}& \textbf{AV}  \\%
\hline
\multirow{3}{*}{Scenario 1}
& Scenario 1 & 0.8470 & 0.6965 & 0.7117& 0.8775 \\
& Scenario 2 &  0.2938 & 0.2955 & 0.2616& 0.3490 \\
& Scenario 3 &  0.1471 & 0.1355 & 0.1410& 0.1528 \\
\hline
\multirow{3}{*}{Scenario 2}
& Scenario 1 &  0.2986 & 0.1918 & 0.1844& 0.3060 \\
& Scenario 2 & 0.7600 & 0.5838 & 0.4987& 0.7418 \\
& Scenario 3 & 0.1504 & 0.1486 & 0.1243 & 0.2049 \\
\hline
\multirow{3}{*}{Scenario 3}
& Scenario 1 & 0.2309 & 0.1479 & 0.1571 & 0.2767 \\
& Scenario 2 & 0.2032 & 0.1229 & 0.1063 & 0.2182 \\
& Scenario 3 & 0.6736 & 0.2240 & 0.3013 & 0.5708 \\
 \hline
All & All & \multirow{2}{*}{0.7702} & \multirow{2}{*}{0.6335} & \multirow{2}{*}{0.6393} & \multirow{2}{*}{0.8412} \\
scenarios &  scenarios& & & &\\
\hline
\end{tabular}
\caption{{\small Test accuracy for teacher models. D: DualCamNet. R: ResNet-50 \cite{He2016}. T: Temporal ResNet-50 \cite{Garcia2018}. AV: AVNet.}}
\label{table:teachers_accuracy}
\end{table}

\textbf{Student Networks.} In order to measure the improvement brought by distillation, we need to look first at the performance of the two proposed student networks when trained only from hard labels only. Column G from Tables~\ref{table:soundnet_accuracy} and~\ref{table:hearnet_accuracy} show the accuracy results for OurSoundNet and HearNet, respectively. It can be observed that both networks perform well, with HearNet achieving a higher accuracy in all scenarios settings.

This result is impressive considering that OurSoundNet was fine-tuned from SoundNet-5 which was trained on the Flickr-SoundNet dataset, while HearNet instead was trained from scratch on our dataset. A reasonable explanation for this is that shallow networks such as HearNet perform better under small data regimes.

\begin{table}[ht]
\centering
\footnotesize
\begin{tabular}{|l|l|c|c|c|}
\hline
\multicolumn{1}{|c|}{\textbf{Train set}} & \multicolumn{1}{c|}{\textbf{Test set}} & \textbf{G} & \textbf{D} & \textbf{R} \\
\hline
\multirow{3}{*}{Scenario 1}
& Scenario 1 & 0.4881 & 0.6071 & 0.5238  \\
& Scenario 2 & 0.4114 & 0.4669 & 0.4378  \\
& Scenario 3 & 0.1958 & 0.2844 & 0.1958  \\
\hline
\multirow{3}{*}{Scenario 2}
& Scenario 1 & 0.4339 & 0.3598 & 0.4220  \\
& Scenario 2 & 0.3333 &   0.3810 & 0.2619  \\
& Scenario 3 & 0.1931 & 0.1799 & 0.1786  \\
\hline
\multirow{3}{*}{Scenario 3}
& Scenario 1 & 0.3796 & 0.4352 & 0.3955  \\
& Scenario 2 & 0.2513 & 0.3386 & 0.2725  \\
& Scenario 3 & 0.3690 & 0.3452 & 0.2619 \\
\hline
\multirow{2}{*}{All scenarios} 
& \multirow{2}{*}{All scenarios} & \multirow{2}{*}{0.4102} & \multirow{2}{*}{0.5299} & \multirow{2}{*}{0.4145} \\
& & & & \\
\hline
\end{tabular}
\caption{{\small Test accuracy for OurSoundNet 
trained with distinct supervisory information. G: Ground truth hard labels. D: DualCamNet soft labels. R: ResNet-50 
soft labels.}}
\label{table:soundnet_accuracy}
\end{table}

\begin{table}[ht]
\centering
\footnotesize
\begin{tabular}{|l|l|c|c|c|}
\hline
\multicolumn{1}{|c|}{\textbf{Train set}} & \multicolumn{1}{c|}{\textbf{Test set}} & \textbf{G} & \textbf{D} & \textbf{R} \\
\hline
\multirow{3}{*}{Scenario 1}
& Scenario 1 & 0.6548 & 0.7857 & 0.7262 \\
& Scenario 2 & 0.4286 & 0.4325 & 0.4960  \\
& Scenario 3 & 0.1627 & 0.1825 & 0.2989 \\ \hline
\multirow{3}{*}{Scenario 2}
& Scenario 1 & 0.4100 & 0.5542 & 0.5106  \\
& Scenario 2 & 0.3214 & 0.2619 & 0.4524 \\
& Scenario 3 & 0.1627 & 0.1825 & 0.1799  \\ \hline
\multirow{3}{*}{Scenario 3}
& Scenario 1 & 0.3307 & 0.3770 & 0.4405  \\
& Scenario 2 & 0.2976 &  0.3056 & 0.2765  \\
& Scenario 3 & 0.5000 & 0.6190 & 0.6071  \\ \hline
\multirow{2}{*}{All scenarios} 
& \multirow{2}{*}{All scenarios} & \multirow{2}{*}{0.6966} & \multirow{2}{*}{0.7009} & \multirow{2}{*}{0.6282} \\
& & & & \\ \hline
\end{tabular}
\caption{{\small Test accuracy for HearNet \cite{Aytar2017} trained with distinct supervisory information. G: Ground truth hard labels. D: DualCamNet soft labels. R: ResNet-50 soft labels.}}
\label{table:hearnet_accuracy}
\end{table}
\textbf{Teacher-Student Networks.}
Finally, we compare the performance of the student networks when trained by distilling knowledge from the teacher networks. These results are shown in columns D and R from Tables~\ref{table:soundnet_accuracy}~and~\ref{table:hearnet_accuracy}.

We observe that whenever we perform training by transferring either from DualCamNet or ResNet-50 using data from scenario 1, we obtain better results and good generalization.
When transferring from DualCamNet, the improvement can be ascribed to the fact that data acquired in the anechoic room is cleaner than in other scenarios.
Similarly, when transferring from ResNet-50, there is little clutter in the scene, allowing the network to easily capture the objects involved in the action execution.

Such ideal conditions do not occur in scenarios 3 and (especially) in scenario 2, which are considerably more acoustically noisy and visually cluttered. In particular the worst case is scenario 2, where echoes are even more disturbing than background noise of scenario 3. In fact, in scenario 3, distilling from DualCamNet improves accuracy in all cases except for the OurSoundNet student tested on scenario 3. We are not able to improve results in all testing scenarios when training on scenario 2 as echoes introduce too much noise. ResNet-50 soft labels, on the contrary, do not always guarantee an increase in accuracy neither in scenario 2 nor in scenario 3.

Overall we thus notice that distillation from DualCamNet provides higher improvement over distillation from ResNet-50, especially when training on all scenarios.
In some exceptional cases, the teacher is able to help the student even though it could not achieve a good accuracy. For instance, ResNet-50 trained on the scenario 3 achieves a 22.4\% test accuracy and HearNet on the same setting reaches 50.0\% but, when transferring from the ResNet-50 to HearNet the accuracy improves up to 60.71\%.


We validated the chosen hyper-parameters, and found $T=1$ and $\lambda=0.5$ to be the best temperature value and imitation parameter, respectively. This means that we keep the teacher predictions unchanged and give them equal importance than to the hard labels. Interestingly our finding about $\lambda$ is consistent with that of \cite{Garcia2018}.

In summary, these results show that knowledge distillation allows learning more robust features given there is not much noise corrupting the data.

\subsection{Acoustic Features Quality Assessment}

Finally, we tested our student networks trained through distillation on a simple classification task on a standard sound benchmark, the DCASE-2018 dataset \cite{Mesaros2018}. Specifically, we performed both k-NN and SVM classification on the features extracted with our distilled student networks to verify whether the learned representations were general enough to perform well in a different audio domain.

Table~\ref{table:tut_accuracy} reports our results in comparison with those obtained from established baselines \cite{Mesaros2018, Liping2018, Golubkov2018} which were trained on DCASE-2018 and that of SoundNet-5 pre-trained on Flickr-SoundNet.

For OurSoundNet, we employed features both from the fc1 and conv4 layers, in order to be comparable with the conv5 and conv4 of the original SoundNet-5. We notice that OurSoundNet/conv4 outperforms OurSoundNet/fc1 by around 14\%. This might be because fc1 has learned feature which are very specific for our dataset. On the other hand, conv4 layer features for both models perform similarly, because they captures less class-specific information, so also OurSoundNet has more general features.

Finally, for HearNet we considered the features from fc1 and some convolutional layers. We observe that lower layer learn features which are more general and thus transfer better to DCASE-2018. This is reasonable since higher layer encode more label-specific information, and this network was trained from scratch on our dataset. Performances of Hearnet lower layers are similar to those of OurSoundNet/fc1, which was pre-trained on Flickr-SoundNet, but was then adapted to our dataset. In conclusion, features learned with our dataset, which comprises 3 hours of videos, transfer reasonably well to DCASE if compared to features learned from the huge Flickr-SoundNet (2M videos).

\begin{table}
\centering
\footnotesize
\begin{tabular}{|l|c|c|c|}
\hline
\multicolumn{1}{|c|}{\textbf{Features}} & \textbf{Training Dataset} & \multicolumn{2}{|c|}{\textbf{Test accuracy }} \\
\hline \hline
Mesaros \textit{et al.} \cite{Mesaros2018}
 & DCASE-2018 &  \multicolumn{2}{|c|}{0.597} \\
Liping \textit{et al.} \cite{Liping2018}
 & DCASE-2018  & \multicolumn{2}{|c|}{0.798} \\
Golubkov \textit{et al.} \cite{Golubkov2018}
 & DCASE-2018 &  \multicolumn{2}{|c|}{0.801}\\
\hline \hline
 \multirow{1}{*}{HearNet/fc1}
 &  Ours & 0.2419 & 0.2609\\
 \multirow{1}{*}{HearNet/conv5}
 &  Ours & 0.2488 & 0.2740\\
 \multirow{1}{*}{HearNet/conv4}
 &  Ours & 0.2631 & 0.2967\\
 \multirow{1}{*}{HearNet/conv3}
 &  Ours & 0.2754 & 0.3100 \\
  \multirow{1}{*}{HearNet/conv2}
 &  Ours & 0.2810 & 0.3403\\
\hline
\multirow{1}{*}{OurSoundNet/fc1}
 &  Flickr-SoundNet+Ours &  0.2746 & 0.3014\\
 \multirow{1}{*}{OurSoundNet/conv4}
 &  Flickr-SoundNet+Ours &  0.4067 & 0.4420 \\
SoundNet-5/conv5
 & Flickr-SoundNet & 0.4180  & 0.4643 \\
SoundNet-5/conv4
 & Flickr-SoundNet &  0.4184 & 0.4275 \\
 \hline

\end{tabular}
\caption{{\small Dataset transfer results for DCASE-2018 \cite{Mesaros2018}. Feature extracted by the models distilled from DualCamNet presented in Section~\ref{section:model_training} are fed into k-NN (\textit{left}) and SVM (\textit{right}) classifiers.
The number of nearest neighbours is validated on the validation set.}} 
\label{table:tut_accuracy}
\end{table}

\section{Conclusions}\label{section:conclusions}
In this work, we investigate whether and how it is possible to transfer knowledge from visual data and spatialized sound, namely, acoustic images, in order to improve audio classification from single microphone. 
To this end, we take advantage of a special sensor, DualCam, an acoustic-optical camera that provides in output audio-visual data synchronized in time and spatially aligned. Using this sensor, we acquired a novel audio-visually indicated action dataset in 3 different scenarios, from which we aim at extracting information useful for audio classification.

The peculiar nature of the generated acoustic images synchronized with optical frames, never studied before, led to the design of deep learning models in the context of the teacher-student paradigm, in order to assess if this information was transferable and indeed useful for single-channel audio classification.
We highlight here that the proposed teacher-student framework is the first able to distill from 2D visual data and acoustic images to a model taking as input a 1D modality, namely, audio signals.

On a set of experiments, in which we learnt from visual data and acoustic images separately, we found out that the distilled models are effective in the audio classification task.
Future work aims at further exploring the capabilities of this sensor for detection, recognition, self-supervised learning, sound source localization and cross-modal retrieval.


{\small
\bibliographystyle{ieee}
\bibliography{perez2019}
}

\newpage
\begin{appendix}
\appendixpage

\section{Data Preparation}\label{section:augmentation}
We implemented all of our networks and our data processing pipeline using TensorFlow. In particular we store our dataset in multiple compressed TFRecord files, each of which contains 1 second of synchronized data from the three modalities, video images, raw audio waveforms, and acoustic images. We use the \texttt{tf.data} API to retrieve this data and compose at runtime variable length sequences. We grouped contiguous TFRecord files into full audio-video sequences and then randomly sampled shorter length sequences, e.g. we compose a full audio-video sequence of 30 seconds and sample from it 10 sequences of 5 seconds.

\section{Dataset Splitting}
In Section 3 of the paper, we mentioned our dataset consists of 378 audio-video sequences from 30 to 60 seconds each. However we did not comment on how it was split for training purposes. Since only a few sequences were longer than 30 seconds, and in order to keep a balanced dataset, we cropped all the sequences up to 30 seconds and assign 80\% of them for training, 10\% for validation and 10\% for test.

Splitting the dataset this way accounts for 302 training sequences, 39 validation sequences, and 37 test sequences. 
We then extracted sequences of the desired length. In case that the required length was 1 second we extracted 30 samples, while in case the required length was 5 seconds we extracted 6 samples. Extracting more samples would result in a high load of data repeated. Finally to keep some consistence across the experiments, we used a fixed seed for random crops extraction and the epoch number as seed for data shuffling.

\section{Hyperparameter Optimization}
In Section 6 of the paper, we presented the obtained experimental results and mentioned that in some cases we used a different learning rate. Basically we considered only two values, \num{1e-3} and \num{1e-4}. Table~\ref{table:learning_rates} shows the values used throughout all the experiments. For all teacher networks we used a learning rate of \num{1e-4} except for \textit{DualCamNet} which required a bigger value. For the student networks (\textit{OursSoundNet} and \textit{HearNet}) we used a mix of both considered values, and almost always the same across all scenarios settings, except for \textit{HearNet} when trained from \textit{DualCamNet} soft labels on first scenario which required a smaller learning rate.
\begin{table}[h]
    \centering
    \footnotesize
    \begin{tabular}{l|c}
        \multicolumn{1}{c|}{\textbf{Network}} & \textbf{Learning rate} \\
        \hline
        \multirow{1}{*}{DualCamNet} & \multirow{1}{*}{\num{1e-3}} \\
        \multirow{1}{*}{ResNet-50} & \multirow{1}{*}{\num{1e-4}} \\
        \multirow{1}{*}{Temporal ResNet-50} & \multirow{1}{*}{\num{1e-4}}\\
        \multirow{1}{*}{AVNet} & \multirow{1}{*}{\num{1e-4}} \\
        \multirow{1}{*}{HearNet (G)} & \multirow{1}{*}{\num{1e-4}} \\
        \multirow{1}{*}{HearNet (D)} & \multirow{1}{*}{\num{1e-3} and \num{1e-4}} \\
        \multirow{1}{*}{HearNet (R)} & \multirow{1}{*}{\num{1e-3} } \\
        \multirow{1}{*}{OurSoundNet (G)} & \multirow{1}{*}{\num{1e-4}} \\
        \multirow{1}{*}{OurSoundNet (D)} & \multirow{1}{*}{\num{1e-3}} \\
        \multirow{1}{*}{OurSoundNet (R)} & \multirow{1}{*}{\num{1e-4}} \\
    \end{tabular}
    \caption{\small Training learning rates. Supervision is indicated as follows: (G): from ground truth hard labels, (D): from DualCamNet soft labels, (R): from ResNet-50 soft labels.}
    \label{table:learning_rates}
\end{table}

It is worth mentioning that in all cases when training our student networks with distillation, we performed hyperparameters optimization using grid search by cross-validation on the held-out validation set. We basically looked at three hyperparameters, learning rate ($lr$), temperature value ($T$), and imitation parameter ($\lambda$). 

Finally, regarding the transfer learning results, also presented in Section 6 of the paper, we validated the considered number of nearest neighbors $k$. We computed accuracy with odd values between 7 and 15 included for validation set, choose on it best $k$ and use that value for the testing accuracy which we report.

\section{Dataset Qualitative Analysis}
\label{section:analysis}
\begin{figure}[h]
\centering
\begin{subfigure}[t]{.32\linewidth}
\includegraphics[width=\linewidth]{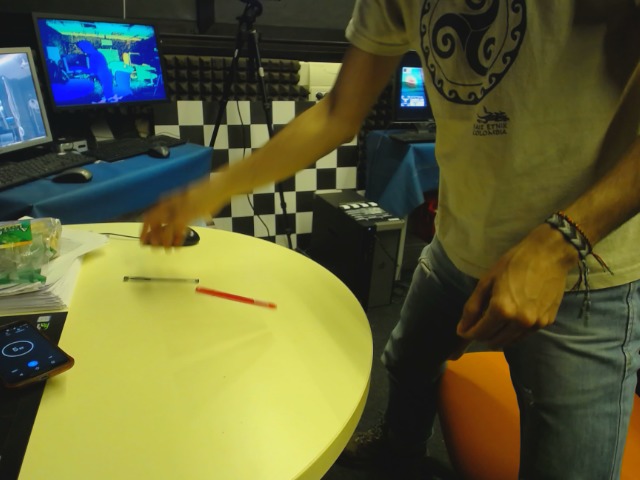}
\includegraphics[width=\linewidth]{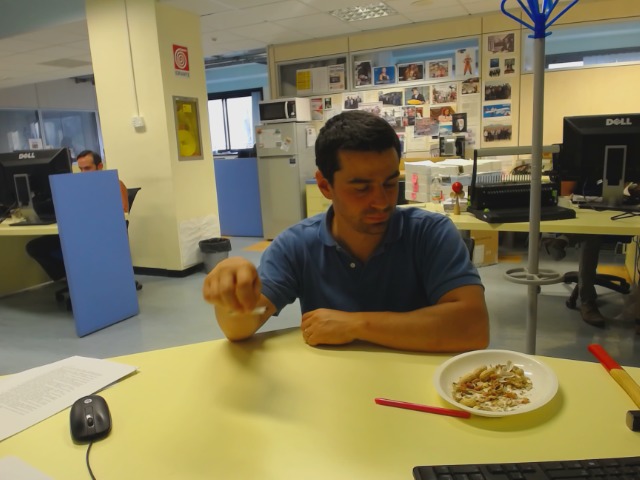}
\includegraphics[width=\linewidth]{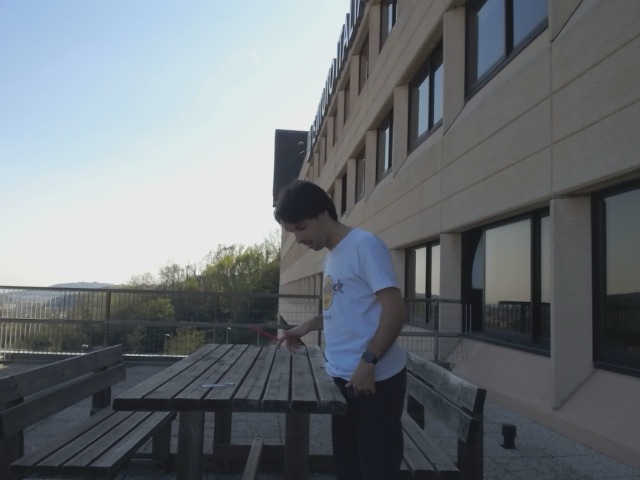}
\caption{\small Stick dropping}\label{fig:1}
\end{subfigure}
\begin{subfigure}[t]{.32\linewidth}
\includegraphics[width=\linewidth]{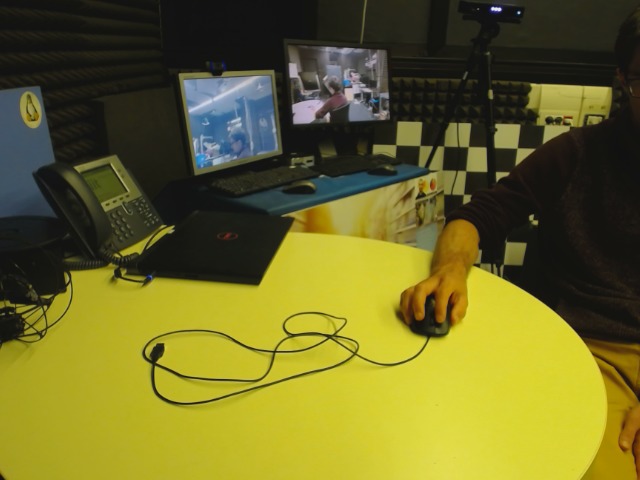}
\includegraphics[width=\linewidth]{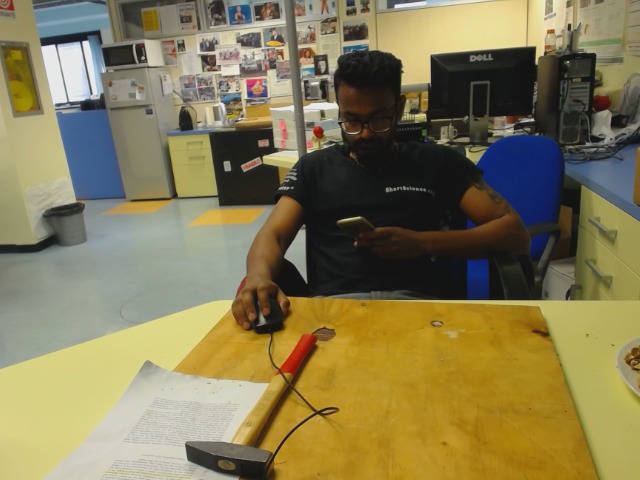}
\includegraphics[width=\linewidth]{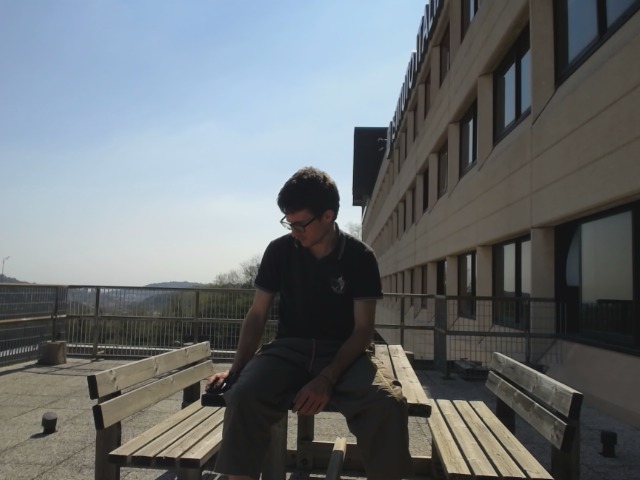}
\caption{\small Clicking}\label{fig:2}
\end{subfigure}
\begin{subfigure}[t]{.32\linewidth}
\includegraphics[width=\linewidth]{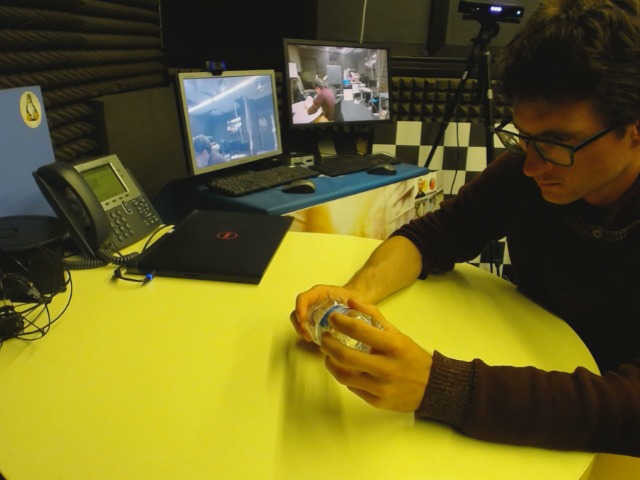}
\includegraphics[width=\linewidth]{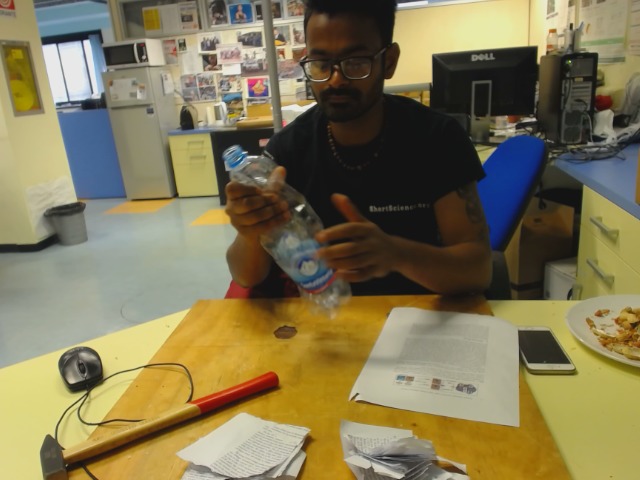}
\includegraphics[width=\linewidth]{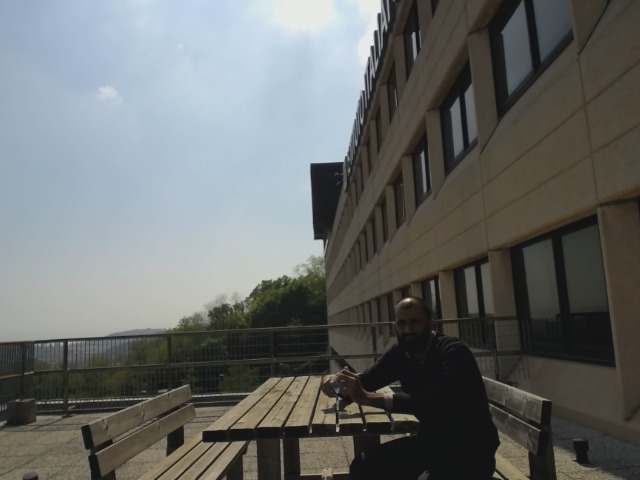}
\caption{\small Plastic crumpling}\label{fig:4}
\end{subfigure}
\caption{\small Comparison of three actions performed on all scenarios. From top to bottom, scenario 1 on the first row, scenario 2 on the second row, and scenario 3 on the third row.}
\label{fig:visual_clutter}
\end{figure}

In this section we provide additional qualitative insights on the proposed dataset, which may clarify some statements made in the paper.
We first illustrate the problem of visual clutter mentioned in Section 6 of the paper. Figure~\ref{fig:visual_clutter} shows three examples of actions performed over all three scenarios with varying conditions of visual clutter.
Comparing scenarios 1 and 3, it can be observed that on the first case the object involved on the action execution is well visible in the foreground, making easier for the visual models to identify the corresponding action.
With scenario 2 the difficulty is that often other people appear on the background or non-related objects are present on the foreground, thus making it harder to identify the action.

\begin{figure}[h]
    \begin{subfigure}[t]{.32\linewidth}
        \includegraphics[width=\linewidth]{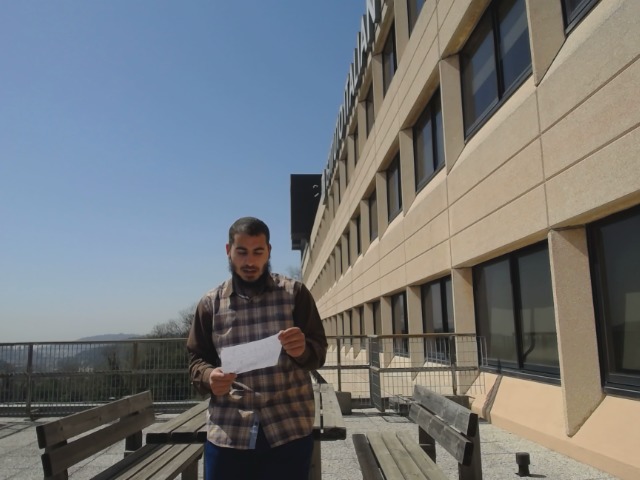}
        \includegraphics[width=\linewidth]{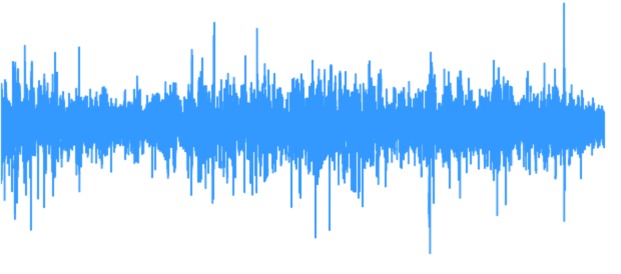}
        \includegraphics[width=\linewidth]{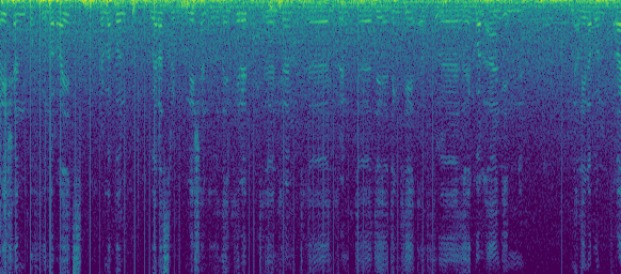}
        \caption{Speaking}\label{fig:8}
    \end{subfigure}
    \begin{subfigure}[t]{.32\linewidth}
        \includegraphics[width=\linewidth]{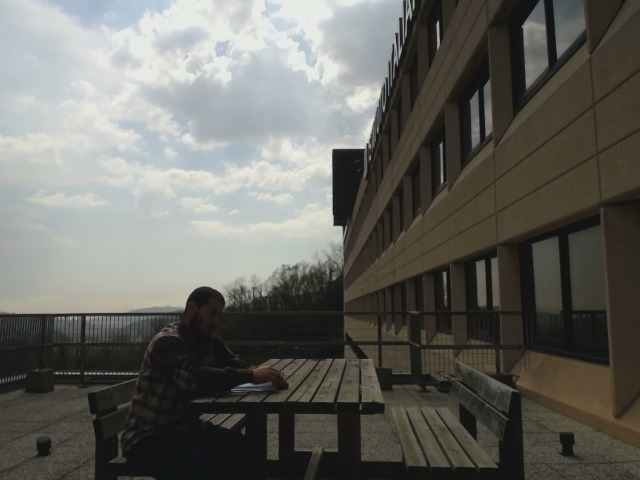}
        \includegraphics[width=\linewidth]{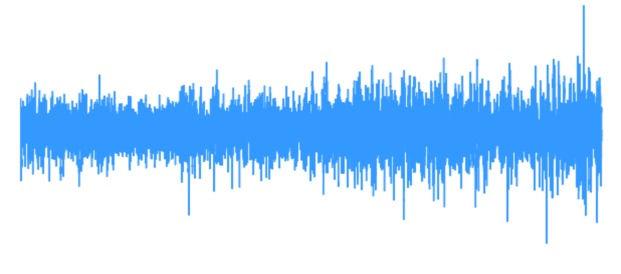}
        \includegraphics[width=\linewidth]{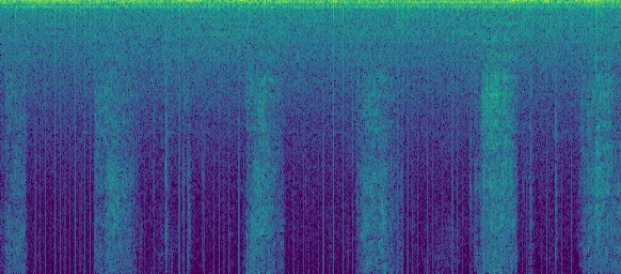}
        \caption{Paper ripping}\label{fig:9}
    \end{subfigure}
        \begin{subfigure}[t]{.32\linewidth}
        \includegraphics[width=\linewidth]{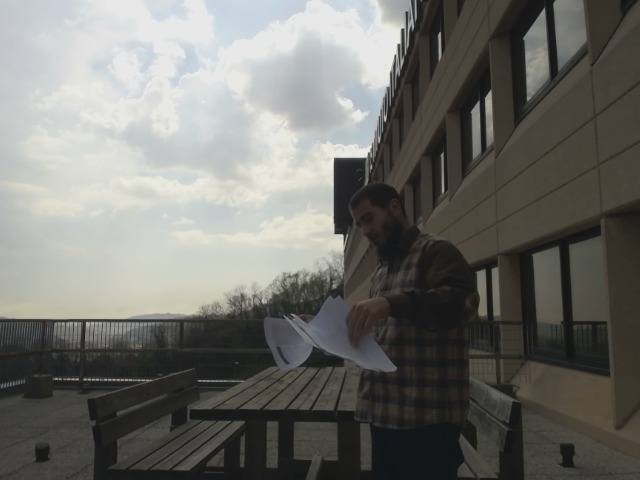}
        \includegraphics[width=\linewidth]{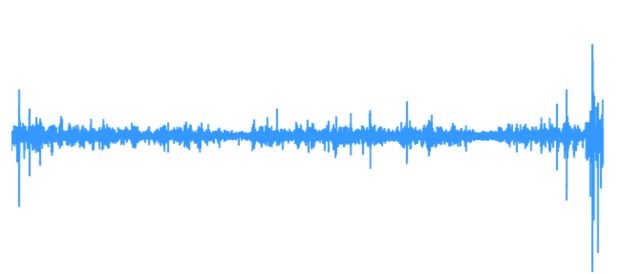}
        \includegraphics[width=\linewidth]{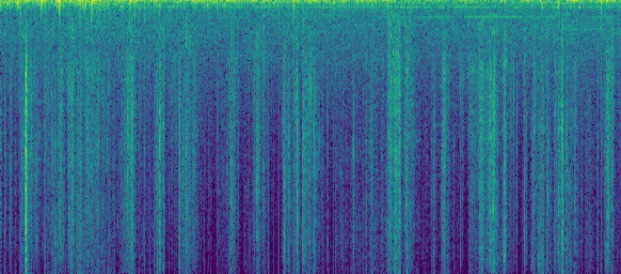}
        \caption{Paper shaking}\label{fig:10}
    \end{subfigure}
    \begin{center}
    \begin{subfigure}[t]{.32\linewidth}
        \includegraphics[width=\linewidth]{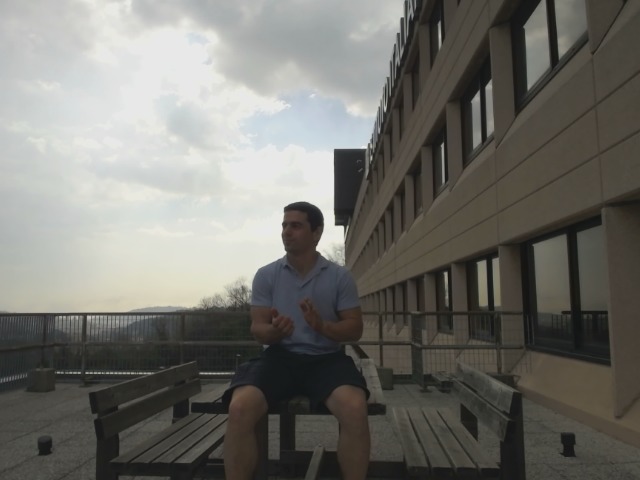}
        \includegraphics[width=\linewidth]{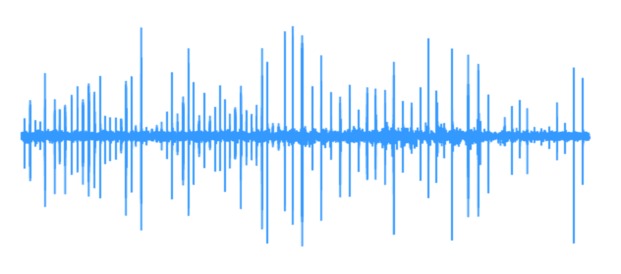}
        \includegraphics[width=\linewidth]{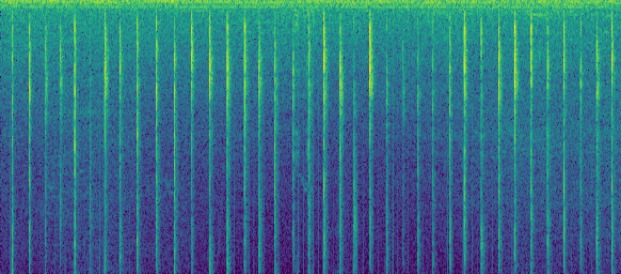}
        \caption{Clapping}\label{fig:11}
    \end{subfigure}
    \begin{subfigure}[t]{.32\linewidth}
        \includegraphics[width=\linewidth]{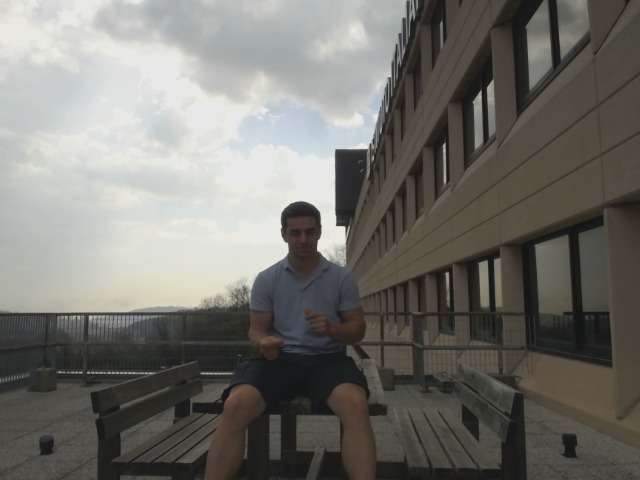}
        \includegraphics[width=\linewidth]{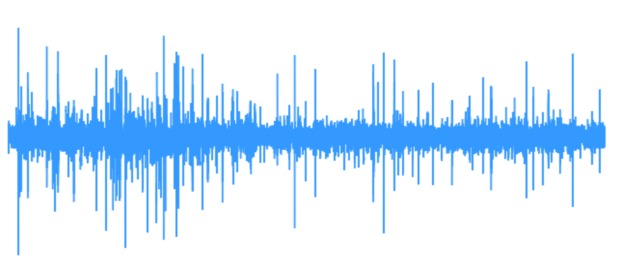}
        \includegraphics[width=\linewidth]{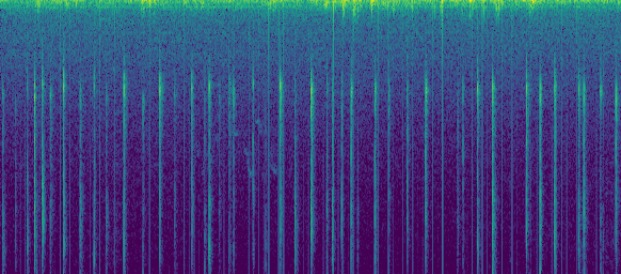}
        \caption{Snapping fingers}\label{fig:12}
    \end{subfigure}
    \begin{subfigure}[t]{.32\linewidth}
        \includegraphics[width=\linewidth]{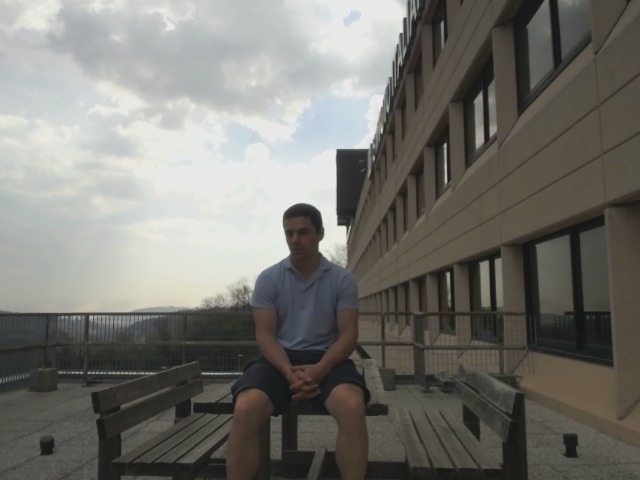}
        \includegraphics[width=\linewidth]{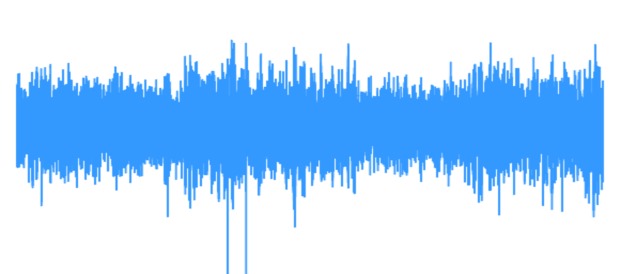}
        \includegraphics[width=\linewidth]{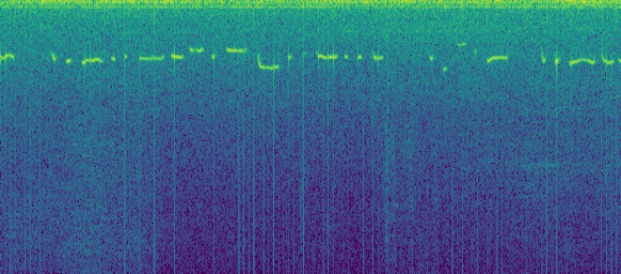}
        \caption{Whistling}\label{fig:12a}
    \end{subfigure}
    \end{center}
    \caption{\small Comparison of six actions visually similar but distinguishable from audio. All six actions where performed on the third scenario corresponding to the terrace.}
    \label{fig:similar_actions}
\end{figure}

A key finding on the paper was that models based on acoustic data achieved better classification results than models based on visual data.
Here we illustrate the difficulty of identifying actions from visual data in contrast to identifying actions from audio data.
Figure~\ref{fig:similar_actions} shows two subjects on the third scenario performing three different actions each.
It can be seen that some actions involving the same subject are visually similar although they depict completely different actions, but they are distinguishable by their acoustic signature.

Looking more closely at Figure~\ref{fig:similar_actions}, it can be seen that some actions have a visually distinguishable pattern. For instance, ``clapping'' and ``snapping fingers'' have a periodic pattern and concentrate on the low frequencies rather than on the high ones. Such patterns are more difficult to grasp from raw waveform. This lead us to think that spectrograms are better audio representations since they summarize the scene acoustic content in a better way when compared to raw waveform. This observation gives some more clues into why \textit{HearNet} performs better than \textit{OurSoundNet} in many cases.

\begin{figure}[h]
    \begin{subfigure}[t]{.32\linewidth}
        \includegraphics[width=\linewidth]{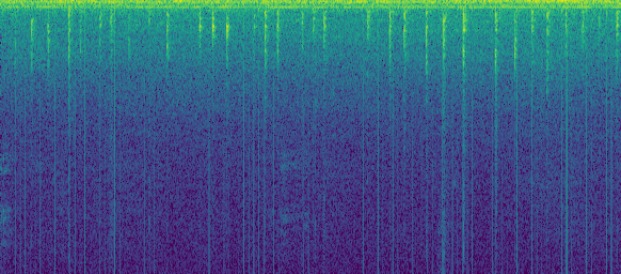}
        \caption{Subject 3}\label{fig:13}
    \end{subfigure}
    \begin{subfigure}[t]{.32\linewidth}
        \includegraphics[width=\linewidth]{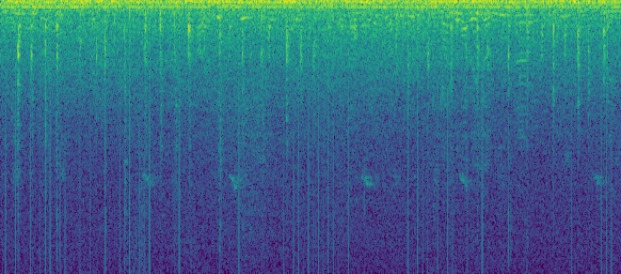}
        \caption{Subject 4}\label{fig:14}
    \end{subfigure}
    \begin{subfigure}[t]{.32\linewidth}
        \includegraphics[width=\linewidth]{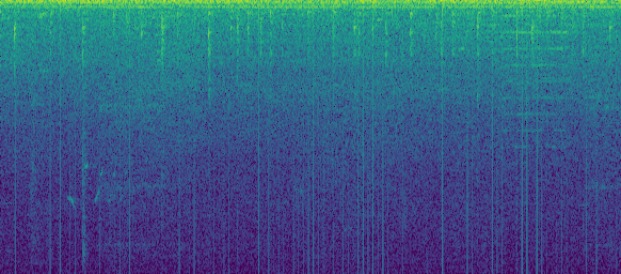}
        \caption{Subject 5}\label{fig:15}
    \end{subfigure} 
    \caption{\small Comparison of the spectrograms for the ``knocking'' action performed by three distinct subjects on the third scenario.}
    \label{fig:subjects_knocking}
\end{figure}

Figure~\ref{fig:subjects_knocking} shows the spectrograms for the same action performed by three different subjects on the third location. There can be seen that the same pattern of multiple events spaced at short time intervals with the energy concentrated on the low frequencies, repeats across different subject executions.

\begin{figure}[h]
    \begin{subfigure}[t]{.32\linewidth}
        \includegraphics[width=\linewidth]{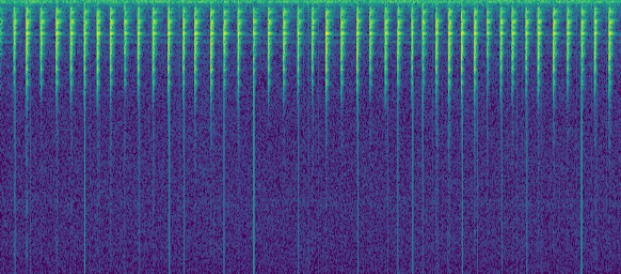}
        \includegraphics[width=\linewidth]{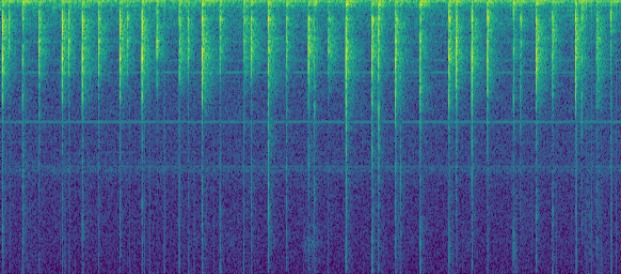}
        \includegraphics[width=\linewidth]{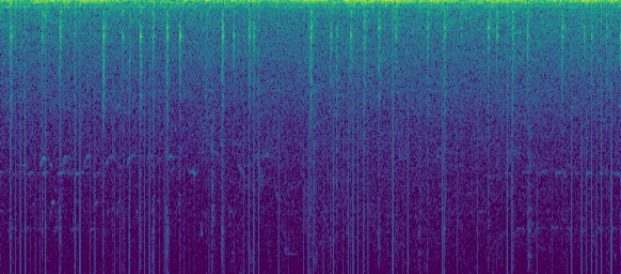}
        \caption{Knocking}\label{fig:22}
    \end{subfigure} 
    \begin{subfigure}[t]{.32\linewidth}
        \includegraphics[width=\linewidth]{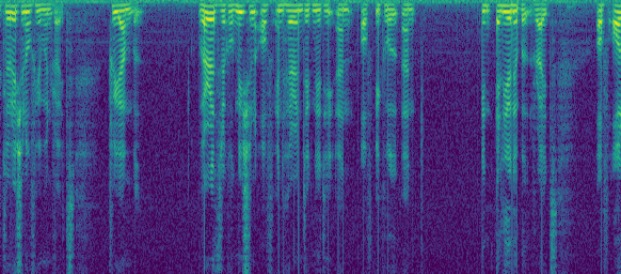}
        \includegraphics[width=\linewidth]{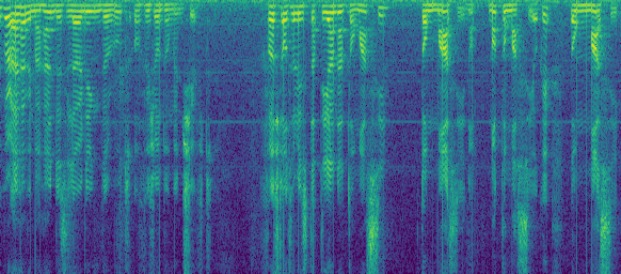}
        \includegraphics[width=\linewidth]{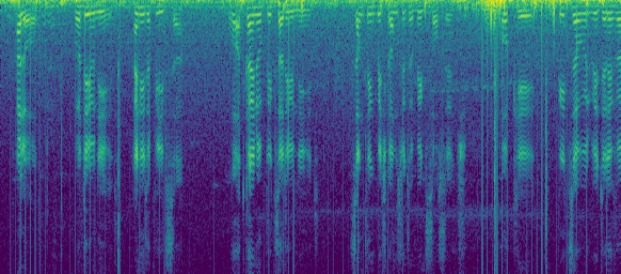}
        \caption{Speaking}\label{fig:23}
    \end{subfigure}
    \begin{subfigure}[t]{.32\linewidth}
        \includegraphics[width=\linewidth]{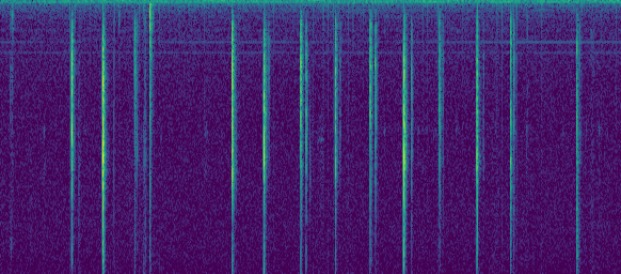}
        \includegraphics[width=\linewidth]{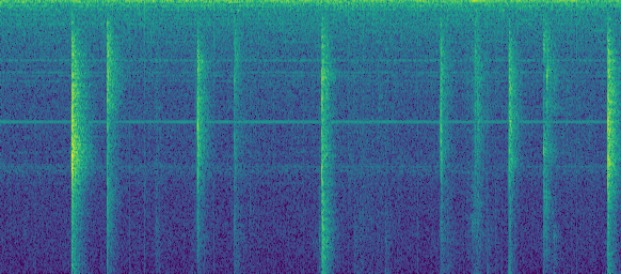}
        \includegraphics[width=\linewidth]{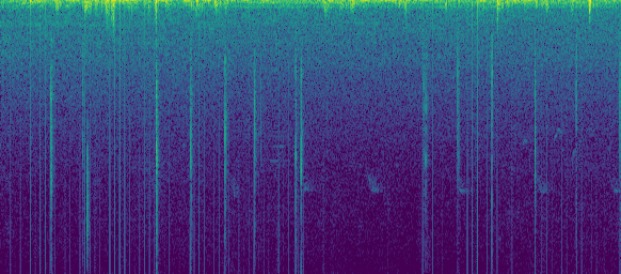}
        \caption{Playing kendama}\label{fig:24}
    \end{subfigure}
    \caption{\small Comparison of the spectrograms of three actions performed by the same subject at the three considered scenarios. From top to bottom, scenario 1 on the first row, scenario 2 on the second row, and scenario 3 on the third row.}
    \label{fig:scenarios}
\end{figure}

Figure~\ref{fig:scenarios} compares the spectrograms of the audios of three different actions performed by the same subject on the three considered scenarios. Here we also see that the audios for the same actions  share a visual pattern when visualized as a spectrogram, even when performed across locations.
Interestingly, the cleanest spectrograms are those from actions performed at first scenario, while for second and third scenarios there are two different kinds of noise. In second scenario the noise is mainly due to indoor echoes, while for third scenario it is due to ambient noise.

\section{Dataset Quantitative Analysis}
We report here the confusion matrices for all the student and teacher models, in order to get a deeper understanding of the dataset's challenges.
\begin{figure}[htbp]
    \centering
    \includegraphics[width=\columnwidth]{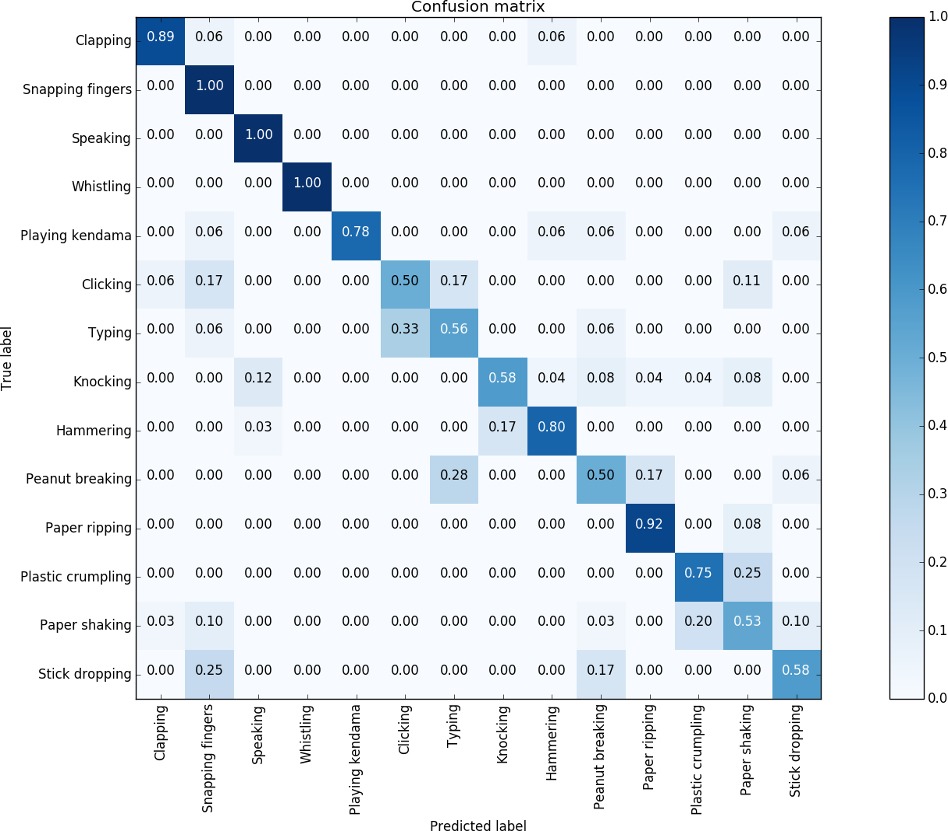}
    \caption{Hearnet trained on all scenarios confusion matrix.}
    \label{fig:hearnet_cm}
\end{figure}

For HearNet (Figure~\ref{fig:hearnet_cm}) we notice that Hammering is often confused with Knocking, Clicking with Typing, Paper shaking with Plastic crumpling. All the three pairs of classes, in fact, are very similar aurally.
\begin{figure}[htbp]
    \centering
    \includegraphics[width=\columnwidth]{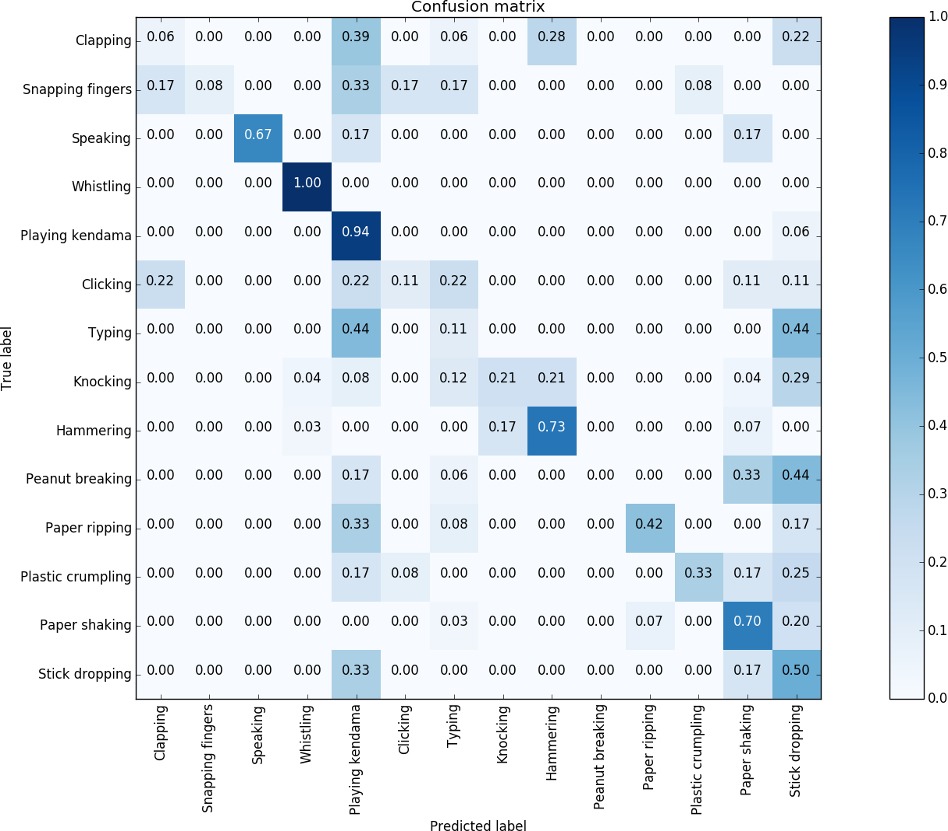}
    \caption{OurSoundNet trained on all scenarios confusion matrix.}
    \label{fig:soundnet_cm}
\end{figure}

Regarding OurSoundNet (Figure~\ref{fig:soundnet_cm}) many classes are confused with Playing kendama and Stick dropping. Hammering and Knocking, Paper shaking and Stick dropping are confused with each other, Peanut breaking is always misclassified, probably because of its feeble audio pattern. As stated before, HearNet superior performance may be ascribed to its more powerful input representation (spectrogram).

\begin{figure}[htbp]
    \centering
    \includegraphics[width=\columnwidth]{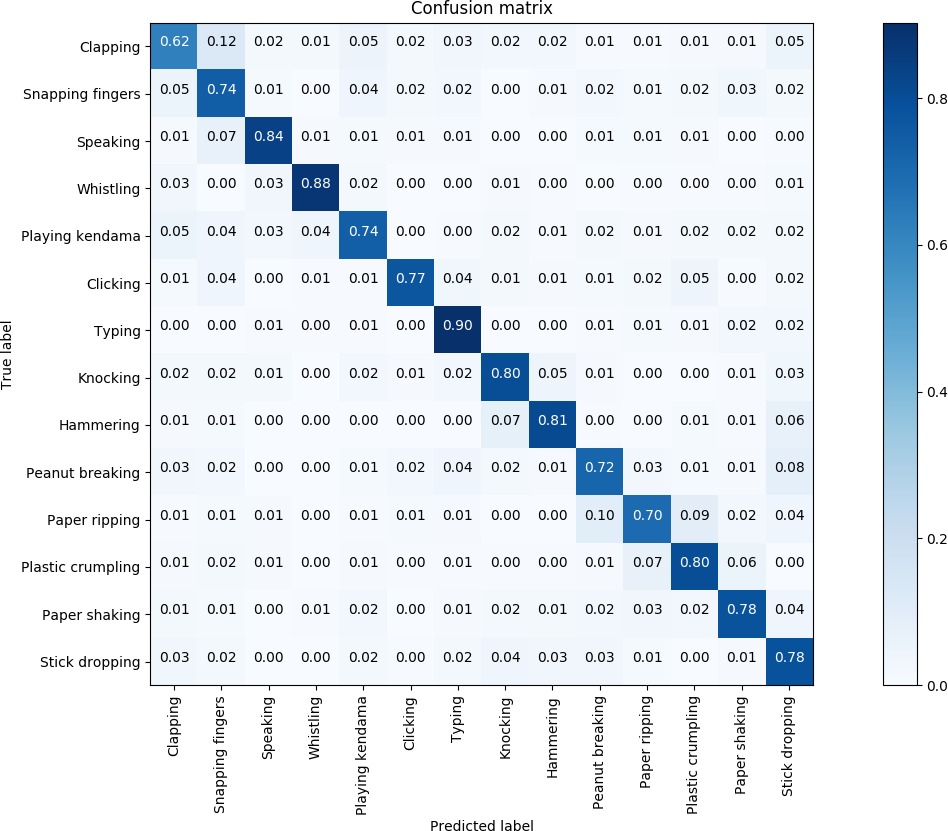}
    \caption{DualCamNet trained on all scenarios confusion matrix.}
    \label{fig:dualcamnet_cm}
\end{figure}

\begin{figure}[htbp]
    \centering
    \includegraphics[width=\columnwidth]{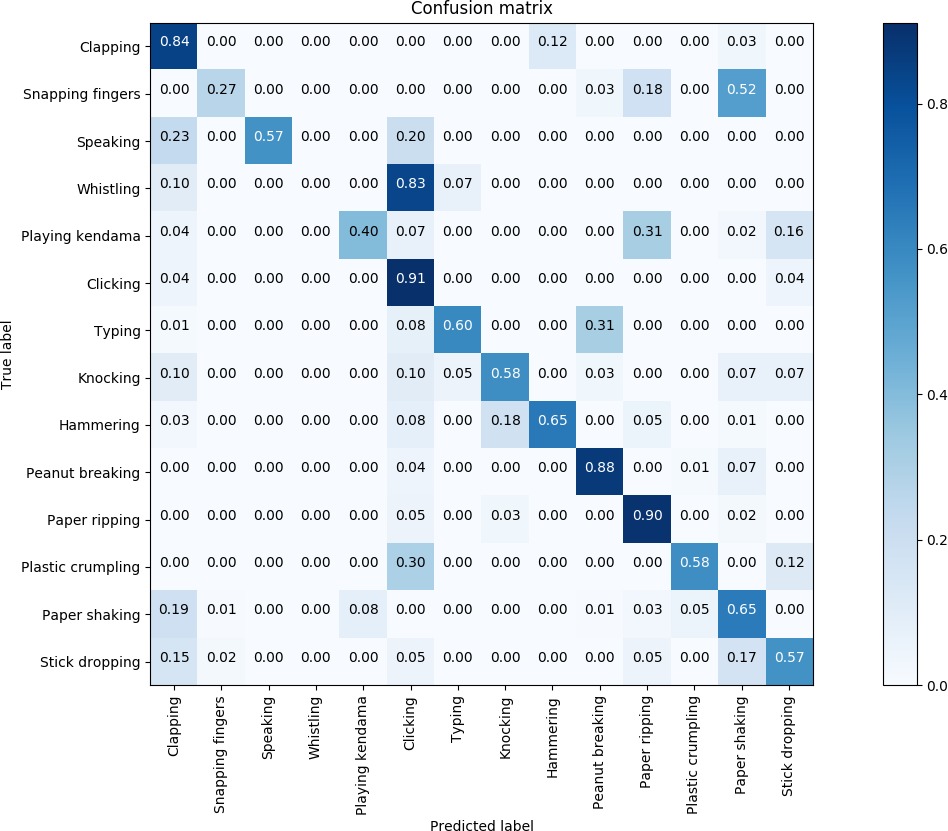}
    \caption{ResNet-50 trained on all scenarios confusion matrix.}
    \label{fig:resnet_cm}
\end{figure}

\begin{figure}[htbp]
    \centering
    \includegraphics[width=\columnwidth]{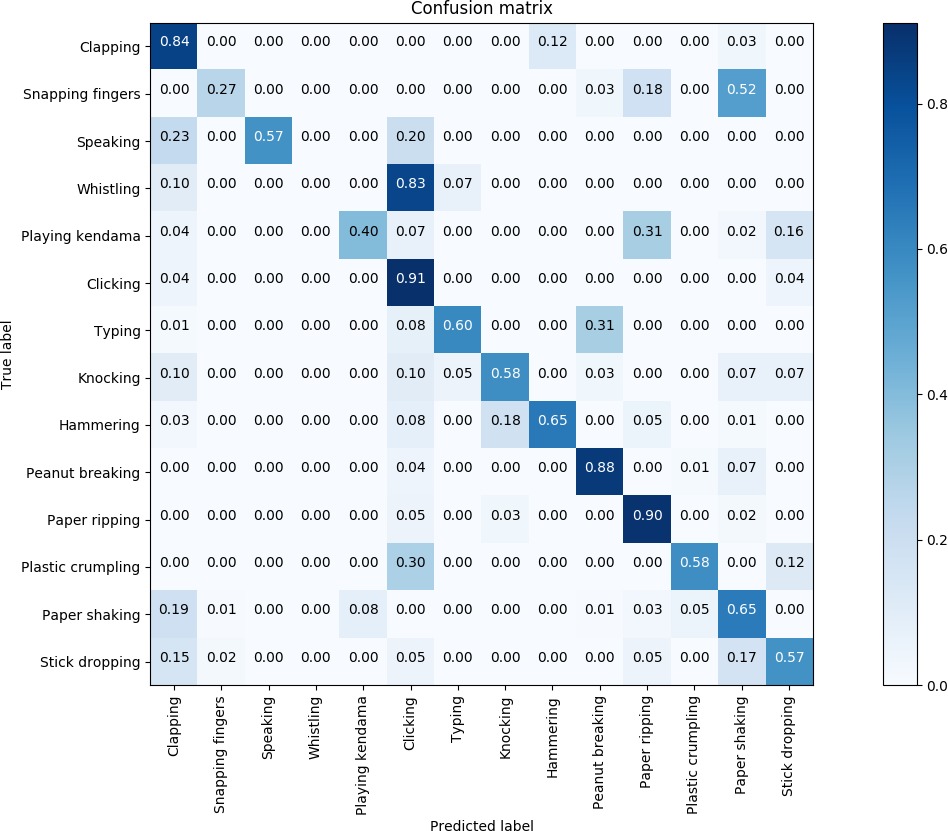}
    \caption{Temporal ResNet-50 trained on all scenarios confusion matrix.}
    \label{fig:tempresnet_cm}
\end{figure}

\begin{figure}[htbp]
    \centering
    \includegraphics[width=\columnwidth]{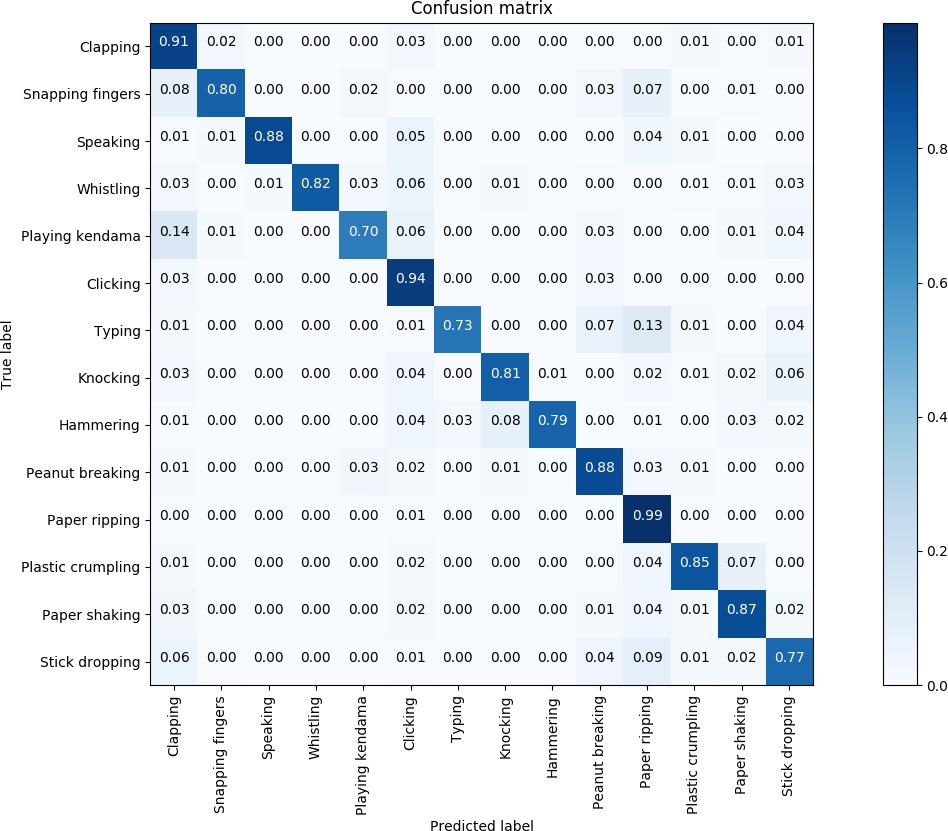}
    \caption{AVNet trained on all scenarios confusion matrix.}
    \label{fig:avenet_cm}
\end{figure}

We now consider the teachers confusion matrices. DualCamNet  (Figure~\ref{fig:dualcamnet_cm}) and AVNet (Figure~\ref{fig:avenet_cm}) confusion matrices have diagonal elements with very high values, indicating high accuracy (they are good teachers indeed).
Temporal ResNet-50 in Figure~\ref{fig:tempresnet_cm} and ResNet-50 in Figure~\ref{fig:resnet_cm} confuse many classes with Clapping and Clicking. Whistling is always misclassified. 
As already certified by higher accuracy, we can conclude that are DualCamNet and AVNet are better teacher.

\begin{figure}[htbp]
    \centering
    \includegraphics[width=\columnwidth]{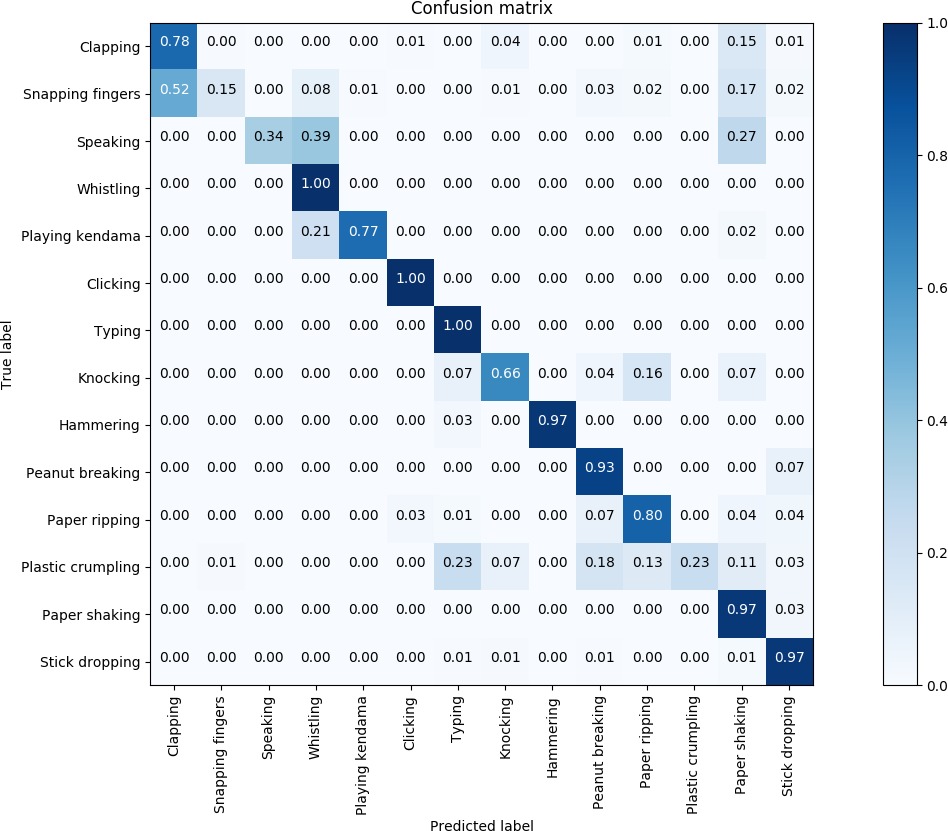}
    \caption{ResNet-50 trained and tested on scenario 1 confusion matrix.}
    \label{fig:resnet_1}
\end{figure}

\begin{figure}[htbp]
    \centering
    \includegraphics[width=\columnwidth]{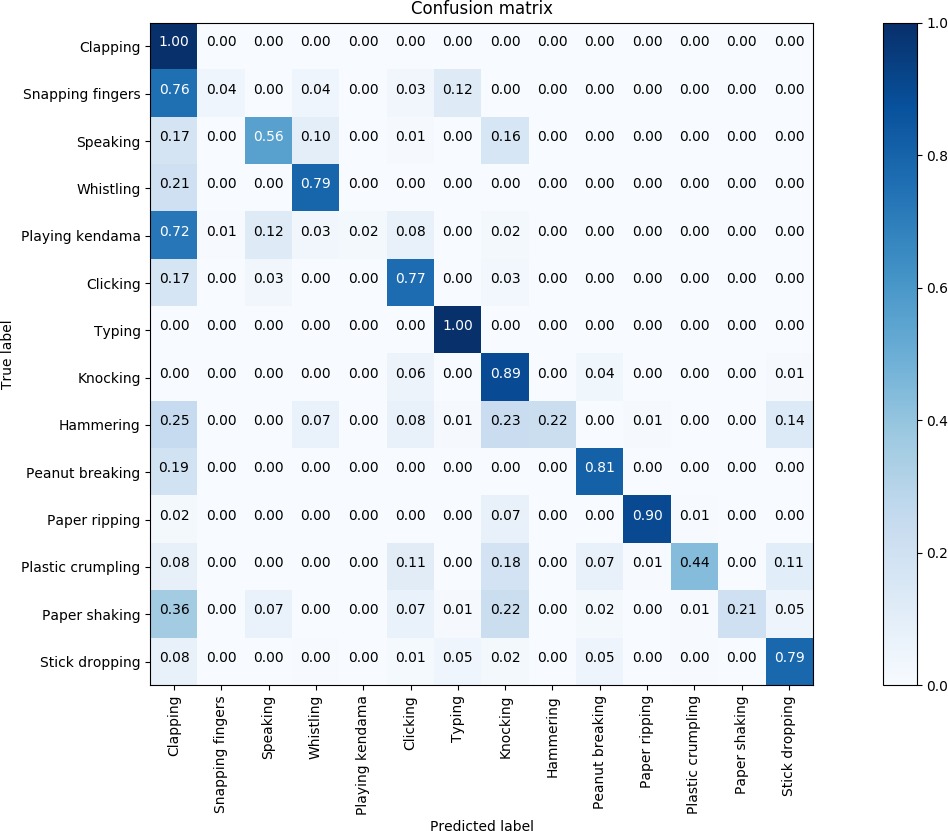}
    \caption{ResNet-50 trained and tested on scenario 2 confusion matrix.}
    \label{fig:resnet_2}
\end{figure}

\begin{figure}[htbp]
    \centering
    \includegraphics[width=\columnwidth]{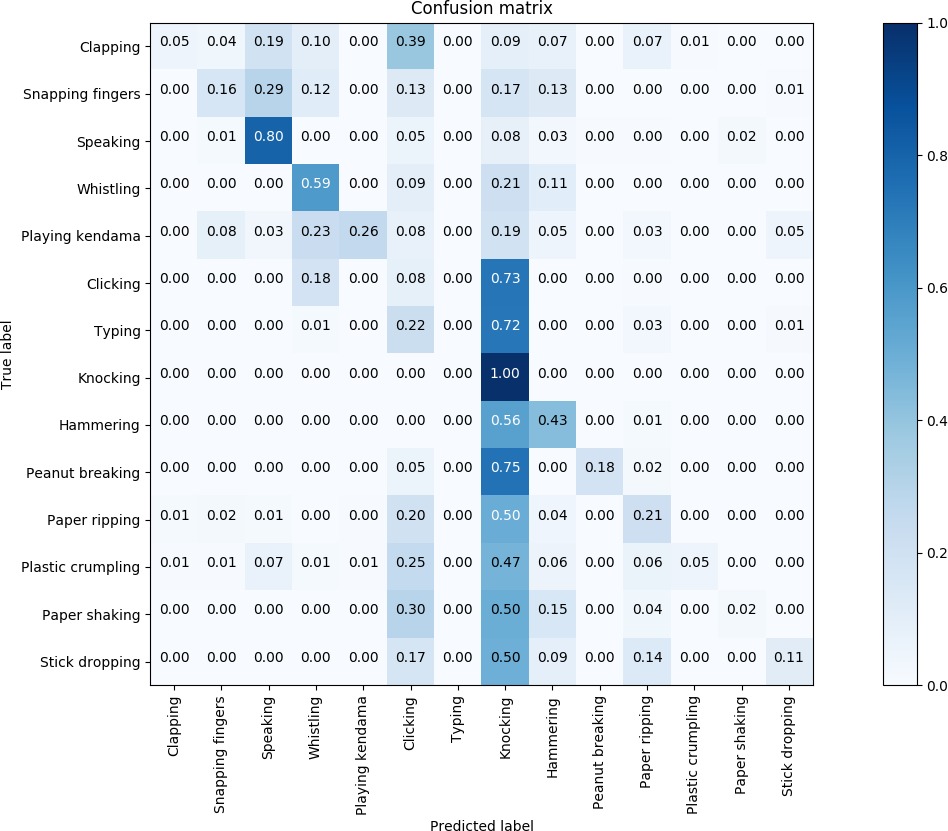}
    \caption{ResNet-50 trained and tested on scenario 3 confusion matrix.}
    \label{fig:resnet_3}
\end{figure}

Finally we can see in detail ResNet-50 confusion matrices when trained and tested on scenario 1 in Figure ~\ref{fig:resnet_1}, scenario 2 in Figure ~\ref{fig:resnet_2} and in scenario 3 in Figure ~\ref{fig:resnet_3}. We notice that when trained and tested on scenario 1, ResNet-50 presents higher accuracies for all classes. In scenario 2 many classes are confused with Clapping, in scenario 3 with Knocking.
In particular, we see in scenario 1 that Snapping fingers, Speaking and Plastic Crumpling are the more difficult to recognize.
In scenario 2 Speaking, Snapping fingers, Playing kendama and Paper shaking have low accuracies.
In scenario 3 many classes have low results, for e.g. Clapping and Snapping fingers. As a matter of fact these classes are visually similar to other ones or sometimes the visual part of the images to recognize the action are occluded or there are other objects  and they can be misunderstood. This confirms the hypothesis made before in Section~\ref{section:analysis} in Figure~\ref{fig:similar_actions} are true.

\section{Reproducibility}

To enable reproducibility of our results and to motivate further research on deep learning for acoustic images, our code\footnote{https://github.com/afperezm/acoustic-images-distillation}, data, and models are publicly available.

\end{appendix}

\end{document}